\documentclass[runningheads]{llncs}

\usepackage{eccv}

\usepackage{eccvabbrv}

\usepackage{graphicx}
\usepackage{booktabs}
\usepackage{mathtools}
\usepackage{caption}
\usepackage{subcaption}
\usepackage{multirow}
\usepackage{siunitx}
\DeclareSIUnit\px{px}
\usepackage{graphicx}
\usepackage{amsmath}
\usepackage{tikz}
\usepackage{pgfplots}
\usepackage{nth}
\pgfplotsset{compat=newest}
\pgfplotsset{every tick label/.append style={font=\tiny}}
\usepgfplotslibrary{groupplots}

\definecolor{RYB1}{RGB}{141, 211, 199}
\definecolor{RYB2}{RGB}{255, 255, 179}
\definecolor{RYB3}{RGB}{190, 186, 218}
\definecolor{RYB4}{RGB}{251, 128, 114}
\definecolor{RYB5}{RGB}{128, 177, 211}
\definecolor{RYB6}{RGB}{253, 180, 98}
\definecolor{RYB7}{RGB}{179, 222, 105}

\pgfplotscreateplotcyclelist{colorbrewer-RYB}{
	{RYB1!50!black,fill=RYB1!50,mark=square*},
	{RYB2!50!black,fill=RYB2!50,mark=*},
	{RYB3!50!black,fill=RYB3!50,mark=pentagon*},
	{RYB4!50!black,fill=RYB4!50,mark=diamond*},
	{RYB5!50!black,fill=RYB5!50,mark=triangle*},
	{RYB6!50!black,fill=RYB6!50,mark=otimes*},
	{RYB7!50!black,fill=RYB7!50},
}

\usepackage{amssymb}
\usepackage{csquotes}
\usepackage{dsfont}
\usepackage{glossaries}
\newacronym{kde}{{KDE}}{kernel density estimation}
\newacronym{gmm}{{GMM}}{Gaussian mixture model}
\newacronym{slam}{{SLAM}}{Simultaneous Localization and Mapping}
\newacronym{sfm}{{SfM}}{Structure-from-Motion}
\newacronym{sift}{{SIFT}}{Scale Invariant Feature Transform}
\newacronym{surf}{{SURF}}{Speeded Up Robust Features}
\newacronym{orb}{{ORB}}{Oriented FAST and rotated BRIEF}
\newacronym{em}{{EM}}{expectation-maximization}
\newacronym{nms}{{NMS}}{non-maximum-suppression}
\newacronym{mnn}{{MNN}}{mutual nearest neighbor}
\usepackage{tikz}
\usepackage{microtype}
\usepackage[export]{adjustbox}
\usepackage{float}
\usepackage{stfloats}
\glsdisablehyper

\usepackage[accsupp]{axessibility}  

\usepackage{hyperref}

\usepackage{orcidlink}

\newcommand{\tbf}[1]{\textbf{#1}}

\DeclareMathOperator*{\argmax}{arg\,max}

\newcommand{\norm}[1]{\left\lVert#1\right\rVert}

\newcommand{\R}{\mathbb{R}}

\definecolor{red}{rgb}{1.00, 0.00, 0.0}

\newcommand{\hinfty}{\raisebox{.2484\height}{$\infty$}}

\begin{document}

\title{GMM-IKRS: Gaussian Mixture Models for Interpretable Keypoint Refinement and Scoring} 
\titlerunning{GMM-IKRS: GMM for Interpretable Keypoint Refinement and Scoring} 

\author{
Emanuele Santellani\inst{1,2}\orcidlink{0009-0001-1301-4423} \and
Martin Zach\inst{1}\orcidlink{0000-0003-1941-875X} \and
Christian Sormann\inst{3}\orcidlink{0000-0002-6824-4007} \and 
Mattia Rossi\inst{3}\orcidlink{0000-0001-5158-2395} \and
Andreas Kuhn\inst{3}\orcidlink{0000-0001-5439-5253} \and
Friedrich Fraundorfer\inst{1}\orcidlink{0000-0002-5805-8892}
}

\authorrunning{E.~Santellani et al.}

\institute{Graz University of Technology, Graz, Austria,
\email{name.surname@tugraz.at} \and
Pro2Future GmbH, Linz, Austria,
\email{name.surname@pro2future.at} \and
Stuttgart Laboratory 1, Sony Semiconductor Solutions Europe, Stuttgart, Germany, \\
\email{name.surname@sony.com}}

\maketitle

\begin{abstract}
The extraction of keypoints in images is at the basis of many computer vision applications, from localization to 3D reconstruction.
Keypoints come with a score permitting to rank them according to their quality.
While learned keypoints often exhibit better properties than handcrafted ones, 
their scores are not easily interpretable,
making it 
virtually impossible
to compare the quality of individual keypoints across methods.
We propose a framework that can refine, and at the same time characterize with an interpretable score, the keypoints extracted by any method.
Our approach leverages a modified robust Gaussian Mixture Model fit designed to both reject non-robust keypoints and refine the remaining ones.
Our score comprises two components: 
one relates to the probability of extracting the same keypoint in an image captured from another viewpoint, 
the other relates to the localization accuracy of the keypoint.
These two interpretable components permit a comparison of individual keypoints extracted across different methods.
Through extensive experiments we demonstrate that, when applied to popular keypoint detectors, 
our framework consistently improves the repeatability of keypoints as well as their performance 
in homography and two/multiple-view pose recovery tasks.
\end{abstract}

\section{Introduction}
\label{sec:intro}

Establishing point correspondences between images is a task that has been of critical importance in the computer vision field since its conception.
Even in the deep learning era,
reliable point-wise matches are still a fundamental requirement for many applications, including 
\gls{sfm}~\cite{schoenberger2016sfm, rome, opensfm},
\gls{slam}~\cite{slamsurvey, orbslam, revo},
visual localization~\cite{visual_localization, outdoorvisuallocalization}
and object tracking~\cite{object_tracking}.
The quality of the established correspondences is crucial for these algorithms when estimating core geometric relationships, such as camera poses and homographies. 
Therefore, there is a high demand for algorithms capable of identifying correspondences,
even in challenging scenarios.

\begin{figure}[t]
\centering
\includegraphics[width=0.8\linewidth]{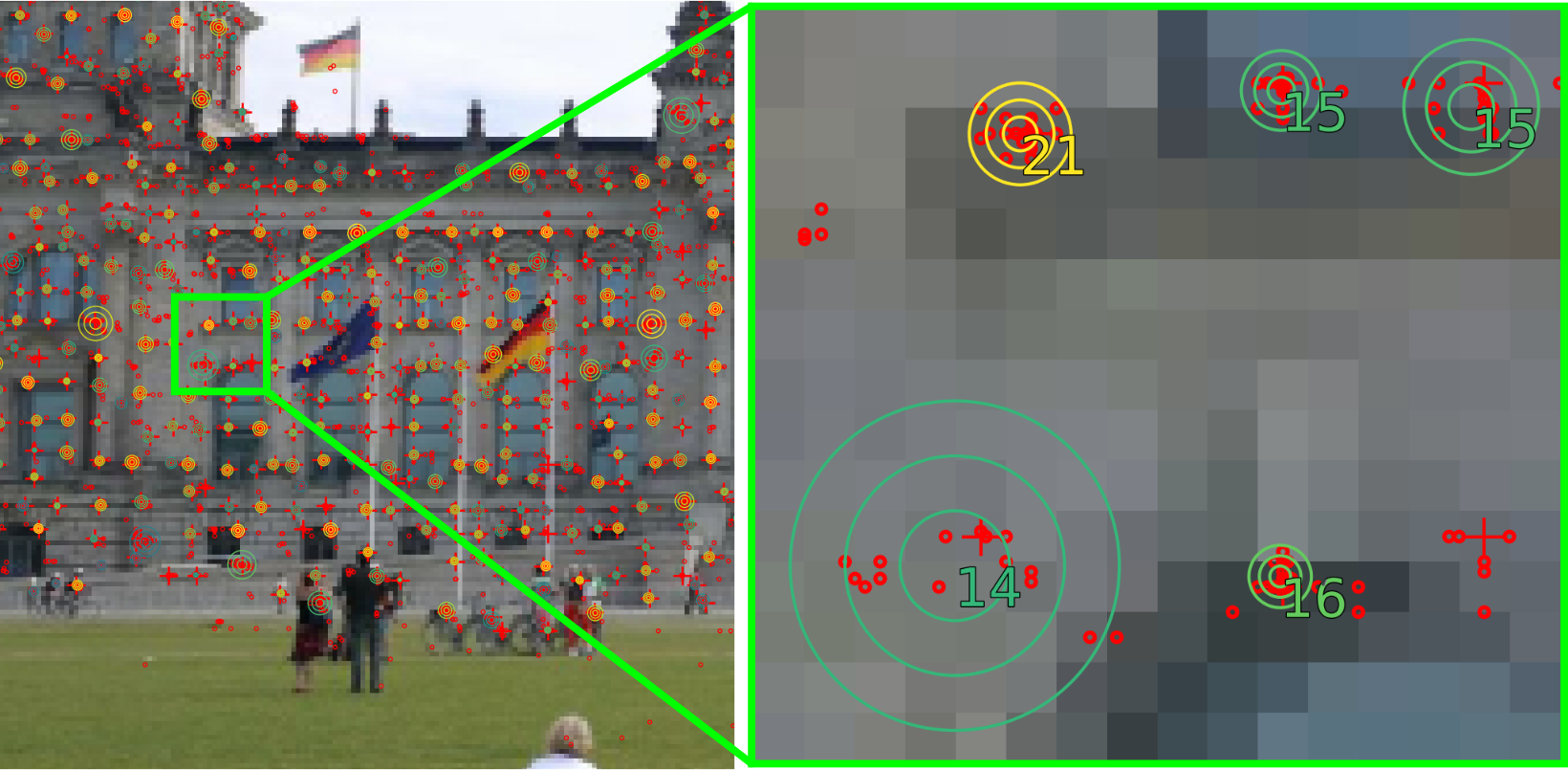}
\caption{
Visualization of the 
input keypoints and their refined positions.
The keypoints from the original image are represented as + in red,
while the red circles represent back-projected keypoints detected in the warped images. 
The Gaussian fit at each keypoint cluster is represented as a set of concentric circles whose spread encodes the variance.
The refined keypoints are the centers of the Gaussians, 
with \textit{robustness} and \textit{deviation} represented by the number next to the Gaussian and by its spread, respectively.
}
\label{fig:teaser}
\end{figure}

Early hand-crafted local feature extractors like the \gls{sift}~\cite{sift}, \gls{surf}~\cite{surf}, and \gls{orb}~\cite{orb}, have been applied in a variety of different applications. 
These methods have established a three step paradigm: keypoint detection, descriptor extraction and pairwise matching. 
In recent years, deep learning has shown great promises for correspondence search, especially in highly challenging scenarios.
This has led to the introduction of many novel paradigms for this task. 
In particular, inherent to their deep learning nature, these architectures oftentimes exhibit less separation between the aforementioned phases.

\begin{figure*}
\centering
\begin{tikzpicture}
  \node[anchor=south west,inner sep=0] at (0,0) {\includegraphics[width=\linewidth]{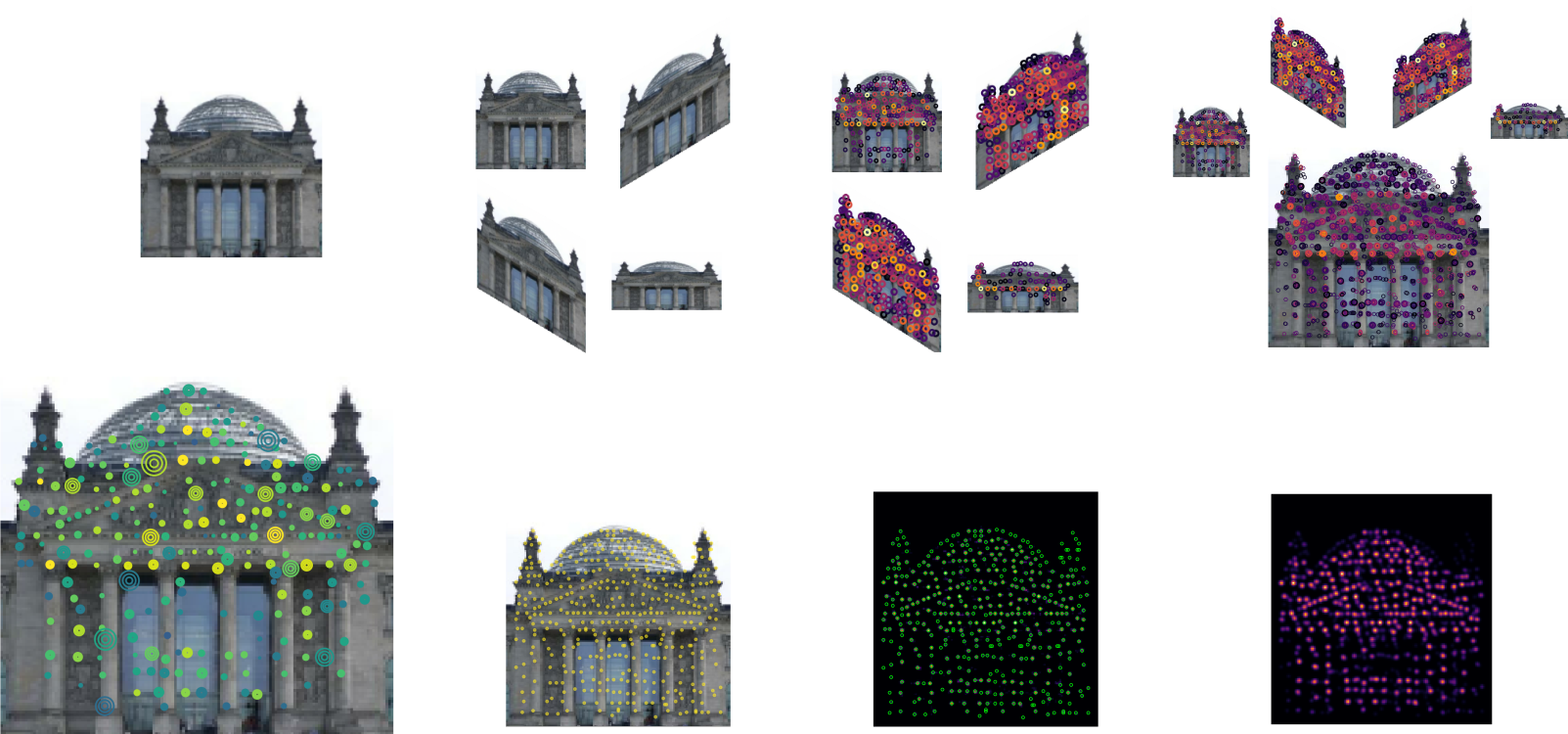}};
  \begin{scope}[shift={(0, 0)}, scale=0.699]
      \node[text opacity=1] at (2.6,5.0) {input image};
      \draw[->, thick] (3.9,6) -- (4.9,6);
      
      \draw[->, thick, red] (6.6,6.7) -- (6.85,6.7);
      \draw[->, thick, red] (6.0,6.2) -- (6.0,5.9);
      \draw[->, thick, red] (6.5,6.2) -- (6.8,5.9);
      \draw[->, thick] (8,6) -- (9,6);
      \node[text opacity=1] at (6.7,4.0) {image warping};
      
      \draw[->, thick] (12.2,6) -- (13.2,6);
      \node[text opacity=1] at (10.7,4.0) {keypoint detection};
      
      \draw[->, thick, red] (13.8,6.0) -- (14.0,5.8);
      \draw[->, thick, red] (14.6,6.9) -- (14.8,6.4);
      \draw[->, thick, red] (16.0,6.9) -- (15.8,6.4);
      \draw[->, thick, red] (17.0,6.5) -- (16.5,6.1);
      \draw[->, thick] (15.4,3.7) -- (15.4,3.3);
      \node[text opacity=1] at (15.4,4.0) {reprojection};
      
      \draw[->, thick] (14,1.4) -- (12.5,1.4);
      \node[text opacity=1] at (15.4,3.0) {KDE};
      
      \draw[->, thick] (9.5,1.4) -- (8.5,1.4);
      \node[text opacity=1] at (11.0,3.0) {NMS};
      
      \draw[->, thick] (5.4,1.4) -- (4.5,1.4);
      \node[text opacity=1] at (6.9,3.0) {GMM initialization};
      
      \node[text opacity=1] at (2.1,4.2) {GMM fit};
  \end{scope}
  
\end{tikzpicture}
\caption{
Sketch of the 
proposed refinement and scoring framework.
A set of image warping
augmentations
are applied to the input image. 
The chosen keypoint detector is applied to all the generated images, and detections in the warped images are projected back to the input image.
The local maxima in the estimated density are used as initialization for a GMM fit. 
After convergence, each Gaussian component represents a refined keypoint characterized by the \emph{robustness} and the \emph{deviation} scores.
This procedure adds additional robust keypoints not detected in the original image.
}
\label{fig:pipeline}
\end{figure*}
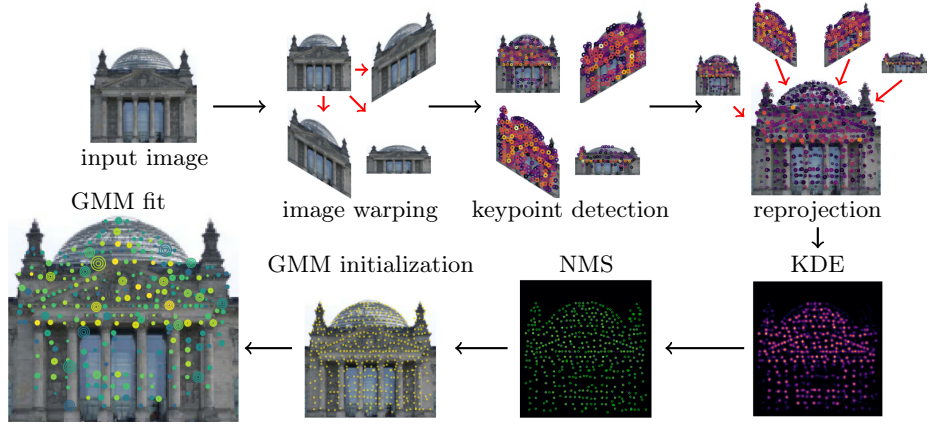

In addition, deep matching algorithms, such as SuperGlue~\cite{superglue}, have emerged. 
These more advanced methods are capable of matching even less discriminative descriptors
by employing a more global form of reasoning.
Despite this advancement,
deep matchers still rely on the input keypoints,
which have now become a primary limiting factor for pose related applications.

Defining what makes a good keypoint is not an easy task,
especially when taking into consideration the specific requirements
of the downstream tasks.
Nevertheless, regardless of how discriminative its surrounding area is, 
a good keypoint must at least be \emph{repeatable}.
In other words, it must be detected again in a different image depicting the same scene. 
Focusing on and expanding the concept of \emph{repeatability},
in this work we introduce a general framework designed to enhance and evaluate keypoints.
Our framework takes an existing keypoint detection method as input,
refines its detections, potentially adding new stable ones, 
and subsequently assigns two distinct scores to each refined keypoint:
the \emph{robustness}, 
which relates to the probability of detecting the keypoint again when the image undergoes viewpoint changes,
and the \emph{deviation}, which quantifies its localization accuracy.
The core of our idea is very simple and strongly relates to the fundamental definition of a good keypoint:
As a first step, we generate multiple versions of the input image by applying some known affine transformations.
We then run the given keypoint detector on each generated image, 
and use the inverse 
transformations to project the detected keypoints back into the original image.
At this point, 
after a first coarse density estimation, 
we use our modified robust \gls{gmm} fit to group the keypoints into clusters.
Our two scores are directly derived from the parameters of each estimated Gaussian.

The proposed refinement,
owing to its compatibility with any keypoint detector
and its linear complexity in the number of images,
presents a valuable option for offline applications requiring robust and well localized keypoints,
such as image-based 3D reconstruction.
Additionally, 
thanks to the interpretability of the two scores,
our method can provide more in-depth insights into the performance of a keypoint detector.
Moreover, our two scores permit to define different keypoint rankings depending on the task at hand.
As an example, more \textit{robust} keypoints might be preferable in visual localization, 
whereas precise robotic applications might prefer low \textit{deviation} ones.

In this work, we present the following contributions:
\begin{itemize}
    \item A novel framework named GMM-IKRS (Gaussian Mixture Models for Interpretable Keypoint Refinement and Scoring) that can refine, 
    and at the same time characterize the keypoints extracted by any detector with two interpretable scores.
    \item A modification of the \gls{em} algorithm designed to make the \gls{gmm} fitting robust to outliers.
    \item Insights into the properties of state-of-the-art learned and hand-crafted detectors through our extensive evaluation.
\end{itemize}

\section{Related works}
The literature on keypoint detectors and local feature extraction methods is large~\cite{lift, surf, fast, aslfeat, d2-net, mdnet, hardnet, brief, brisk, l2-net, lfnet, mser, orb}.
In this section, we focus on those methods that closely relate to our proposed framework 
and refer the reader to \cite{localfeaturesurvey, slamsurvey} for a more extensive overview.

In a similar fashion to the first step of our pipeline,
ASIFT~\cite{asift} runs the SIFT~\cite{sift} feature extractor (keypoint + descriptors)
on multiple warped versions of the original image.
Following this initial step,
the descriptors extracted from each warped image need to be matched across all possible warped image pairs,
which substantially increase the computational cost of the matching phase.
In contrast, 
our framework only extract descriptors from the original image and leaves the matching phase
unaltered.
In addition, ASIFT directly relies on the set of keypoints extracted from the different warpings and does not apply any aggregation; 
instead, our clustering scheme evaluates and refines this initial set to obtain better localized keypoints 
and permits the estimation of our two scores.

The deep method 
SuperPoint~\cite{superpoint} 
also employs a warping augmentation process, 
denoted homographic adaption,
to generate heatmaps utilized in a second stage of the training.
Specifically, random homography warpings are applied to the input image and subsequently processed by the network.
An aggregated heatmap is then computed warping back all the network outputs.
The initial step of our pipeline implements a similar warping scheme, 
however, our framework operates directly on the keypoints, rather than on the heatmaps.
Furthermore, after re-projecting all the keypoints into the original image,
our method applies a robust global clustering algorithm to produce a set of
refined and well characterized keypoints.

After several successful works on joint keypoint and descriptor learning~\cite{d2-net, r2d2, mdnet}, 
promising deep matching algorithms like SuperGlue~\cite{superglue} and LightGlue~\cite{lightglue}
have led to a shift in focus within the research community.
These deep matchers can cope with less reliable keypoint descriptors,
sparkling renewed interest in methods focusing mainly on keypoint detection~\cite{ness, s-trek, keynet}.
The recent work NeSS-ST~\cite{ness} proposes a stability score against 
viewpoint transformation for keypoints detected by the Shi-Tomasi~\cite{shitomasi} detector.
This method assesses the stability of a specified keypoint by firstly applying random homography warpings to a patch centered at the keypoint.
It then picks the highest Shi-Tomasi~\cite{shitomasi} response for each patch
and computes the sample covariance with respect to the original keypoint location.
In contrast,
our method does not require to define any local patch size to compute the scores, 
as our transformations are applied to the whole image.
Moreover, our framework jointly assesses,
in addition to the keypoint localization accuracy,
its robustness conditioned on the detector in analysis. 
Lastly, as opposed to NeSS-ST~\cite{ness},
our framework does not incorporate any learned component and can be applied to arbitrary keypoint detectors.

\noindent
\begin{figure}[t]
\begin{minipage}{0.47\textwidth}
\centering
\includegraphics[width=0.8\linewidth]{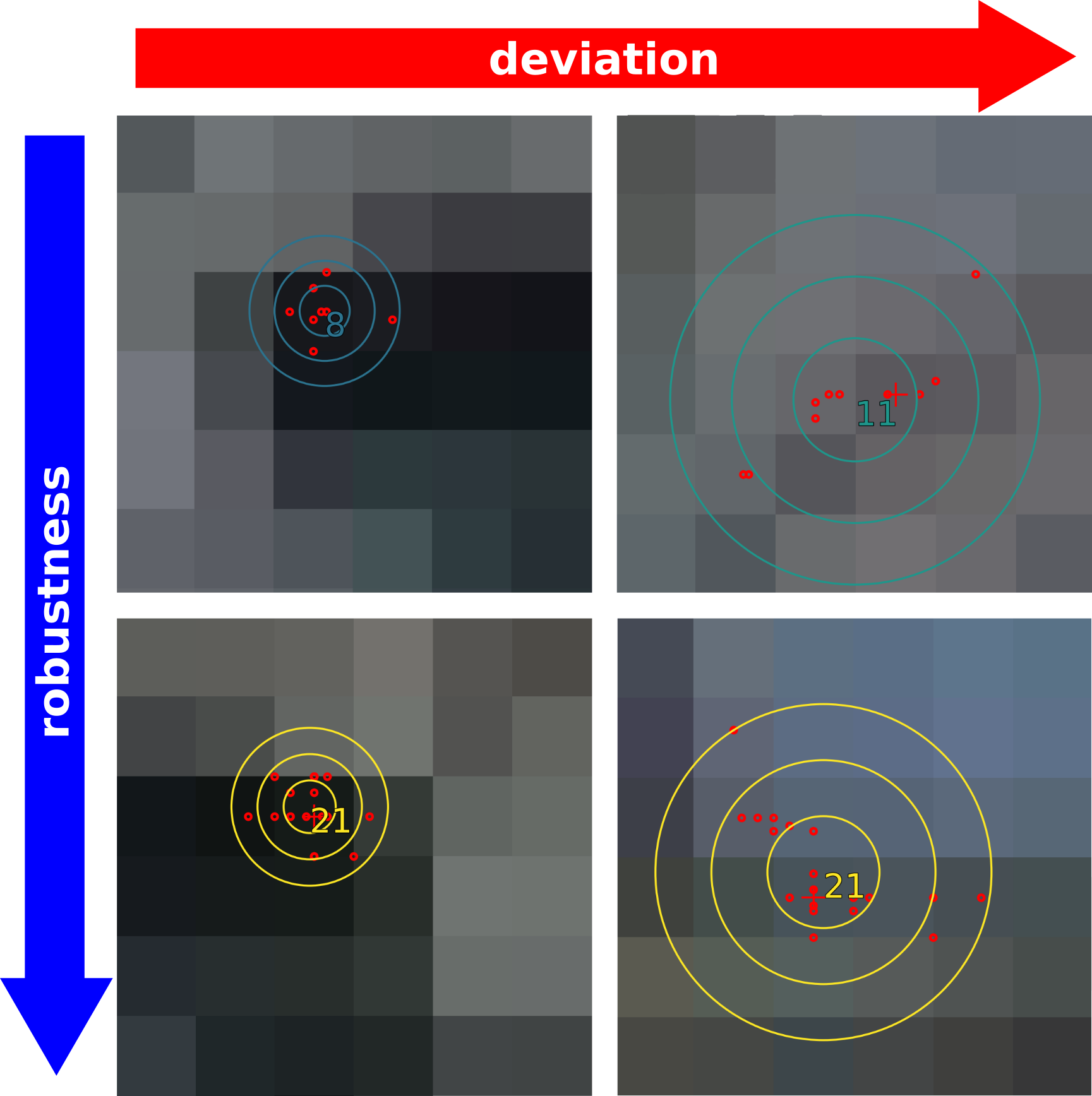}
\captionof{figure}{Keypoint clusters and their two scores: \emph{robustness} and \emph{deviation}. 
\emph{Robustness} measures the likelihood of detecting the keypoint again, while
\emph{deviation} measures its localization accuracy.
A desirable keypoint has high \emph{robustness} and low \emph{deviation}, as in the bottom-left square.
}
\label{fig:cluster_types}
\end{minipage}
\hfill
\begin{minipage}{0.47\textwidth}
     \centering
     \newcommand{\mysize}{0.15\linewidth}
     \centering
     \begin{minipage}[b]{\mysize}
         \includegraphics[width=\linewidth]{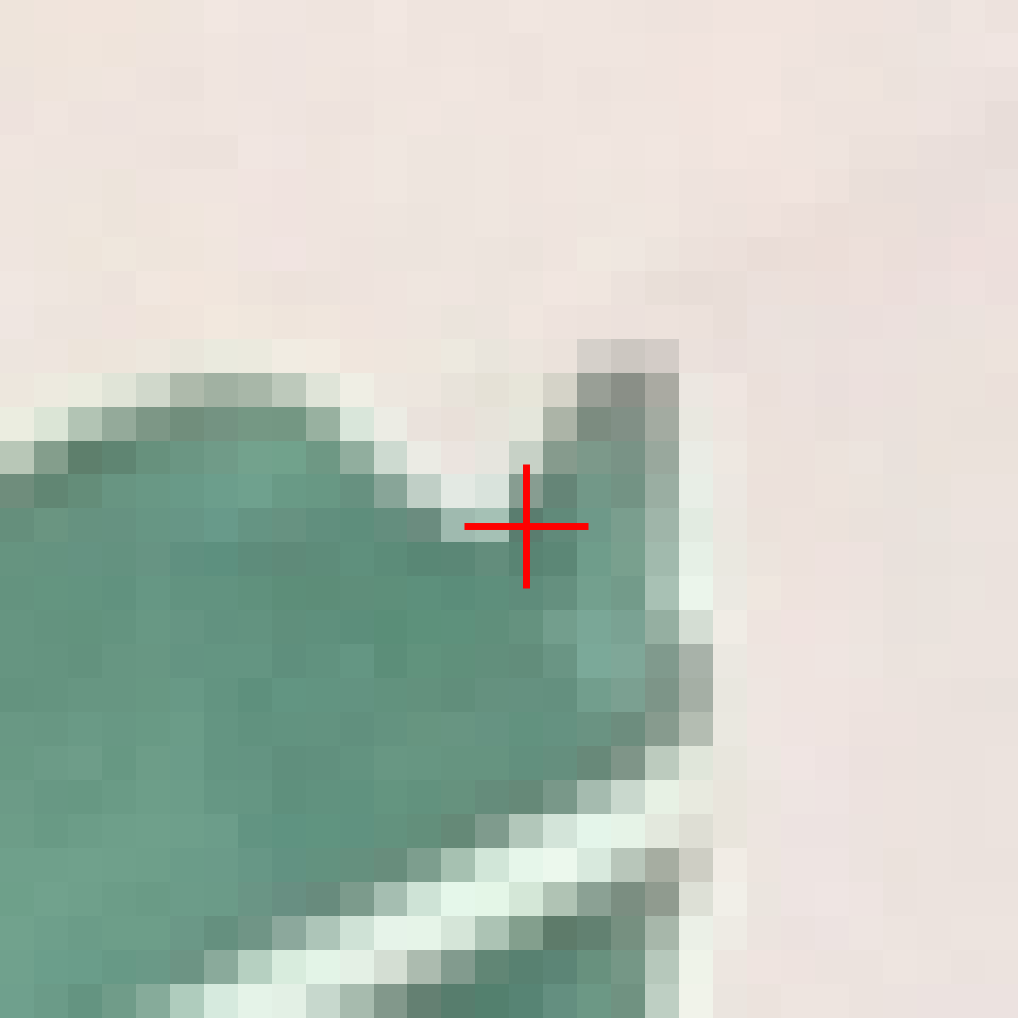}
     \end{minipage}
     \begin{minipage}[b]{\mysize}
         \includegraphics[width=\linewidth]{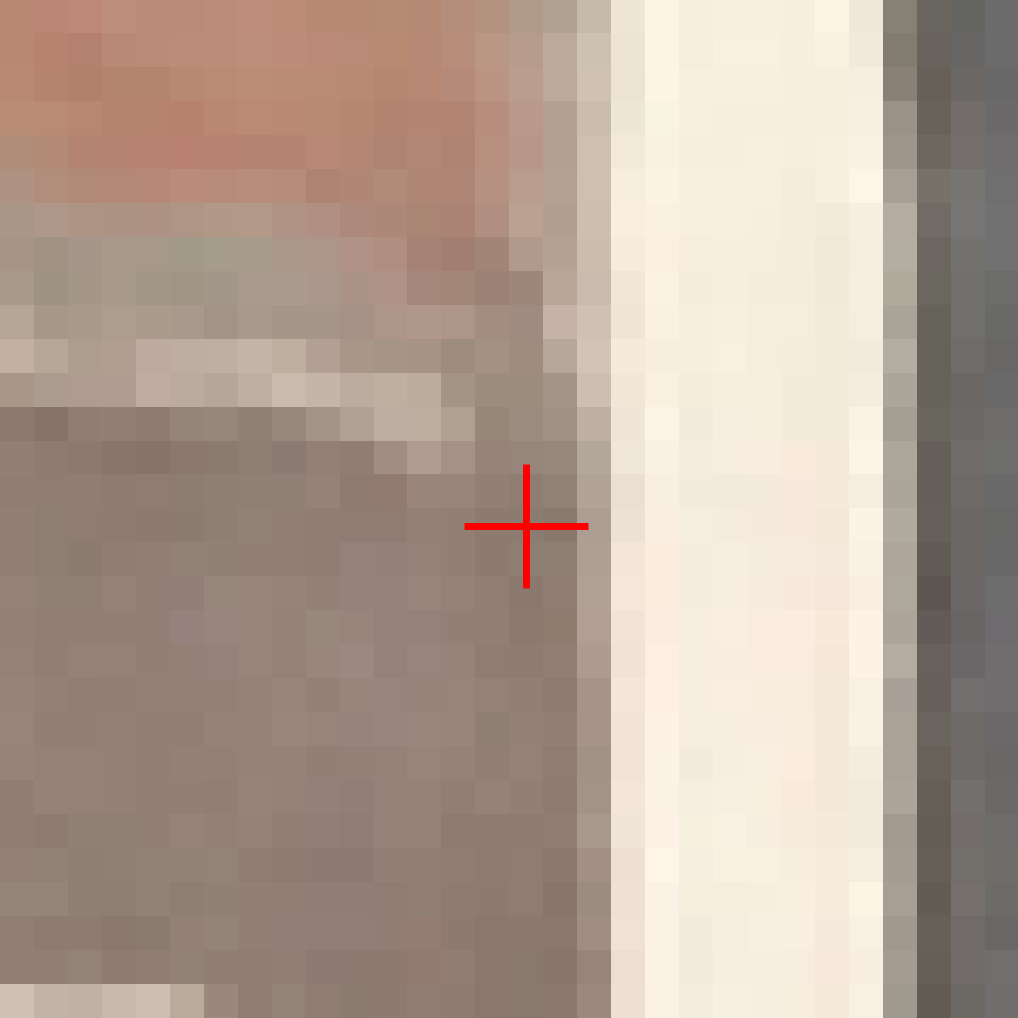}
     \end{minipage}
     \begin{minipage}[b]{\mysize}
         \includegraphics[width=\linewidth]{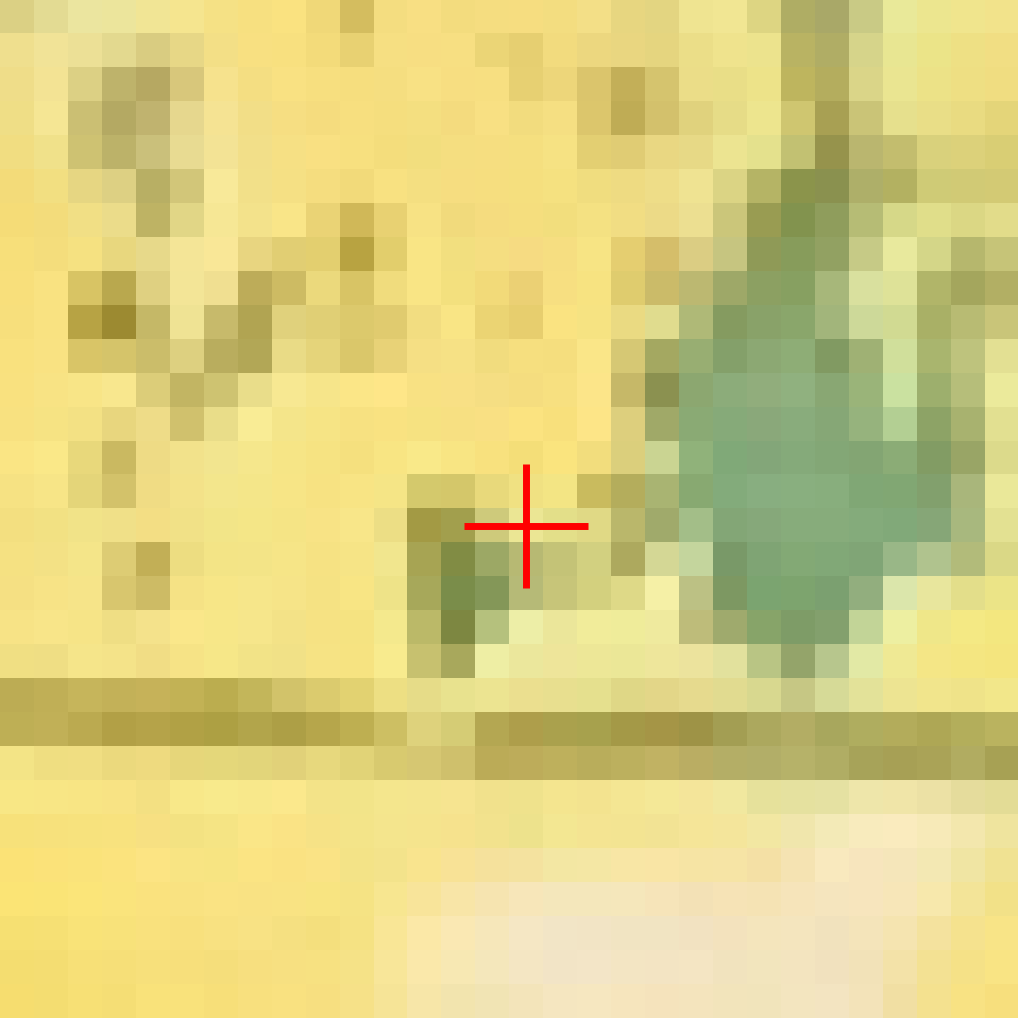}
     \end{minipage}
     \begin{minipage}[b]{\mysize}
         \includegraphics[width=\linewidth]{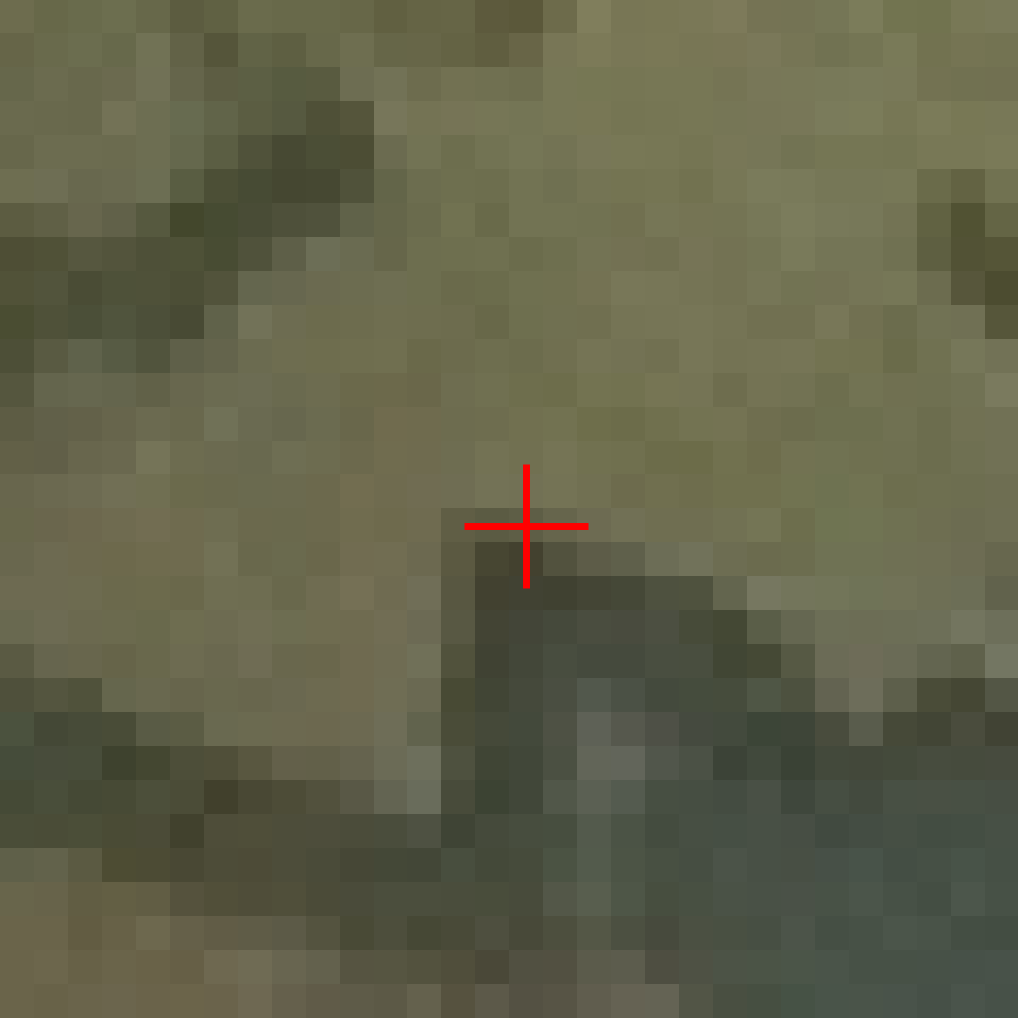}
     \end{minipage}
     \begin{minipage}[b]{\mysize}
         \includegraphics[width=\linewidth]{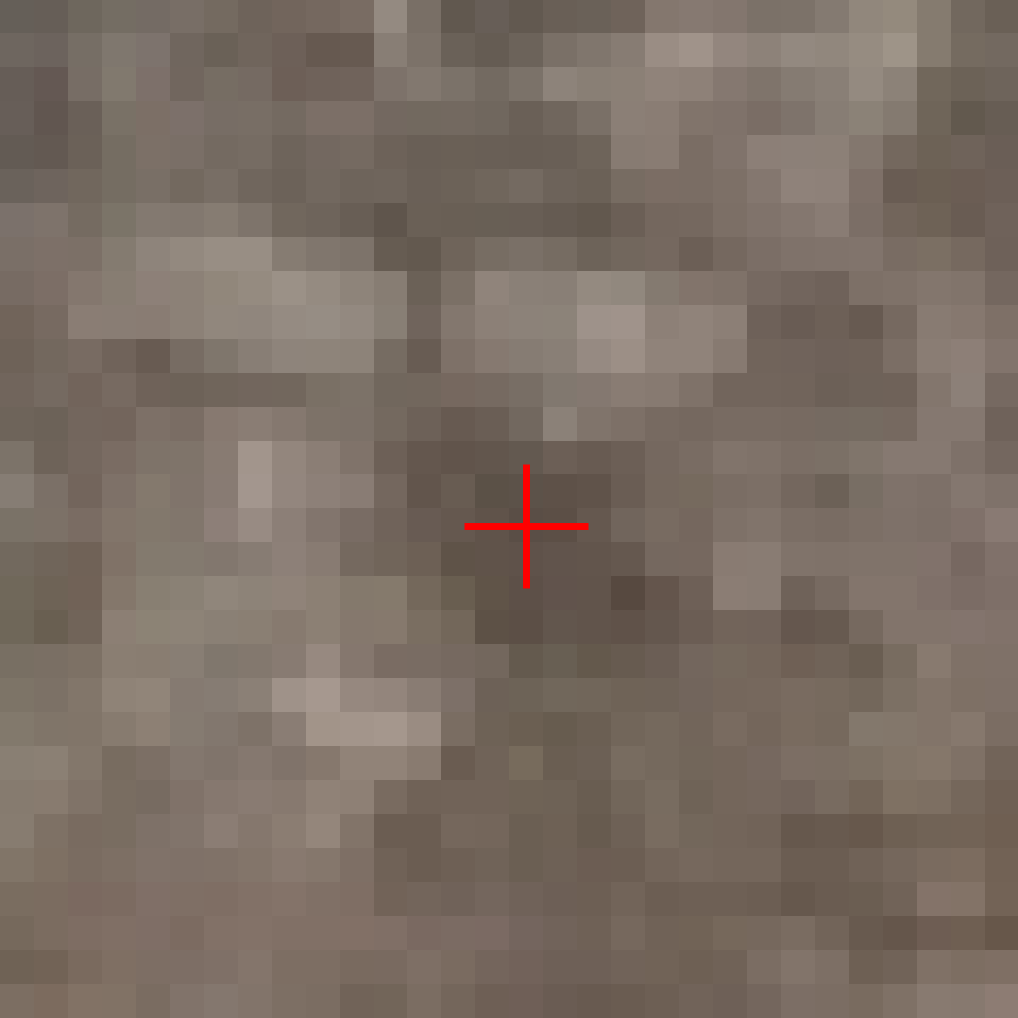}
     \end{minipage}
     \begin{minipage}[b]{\mysize}
         \includegraphics[width=\linewidth]{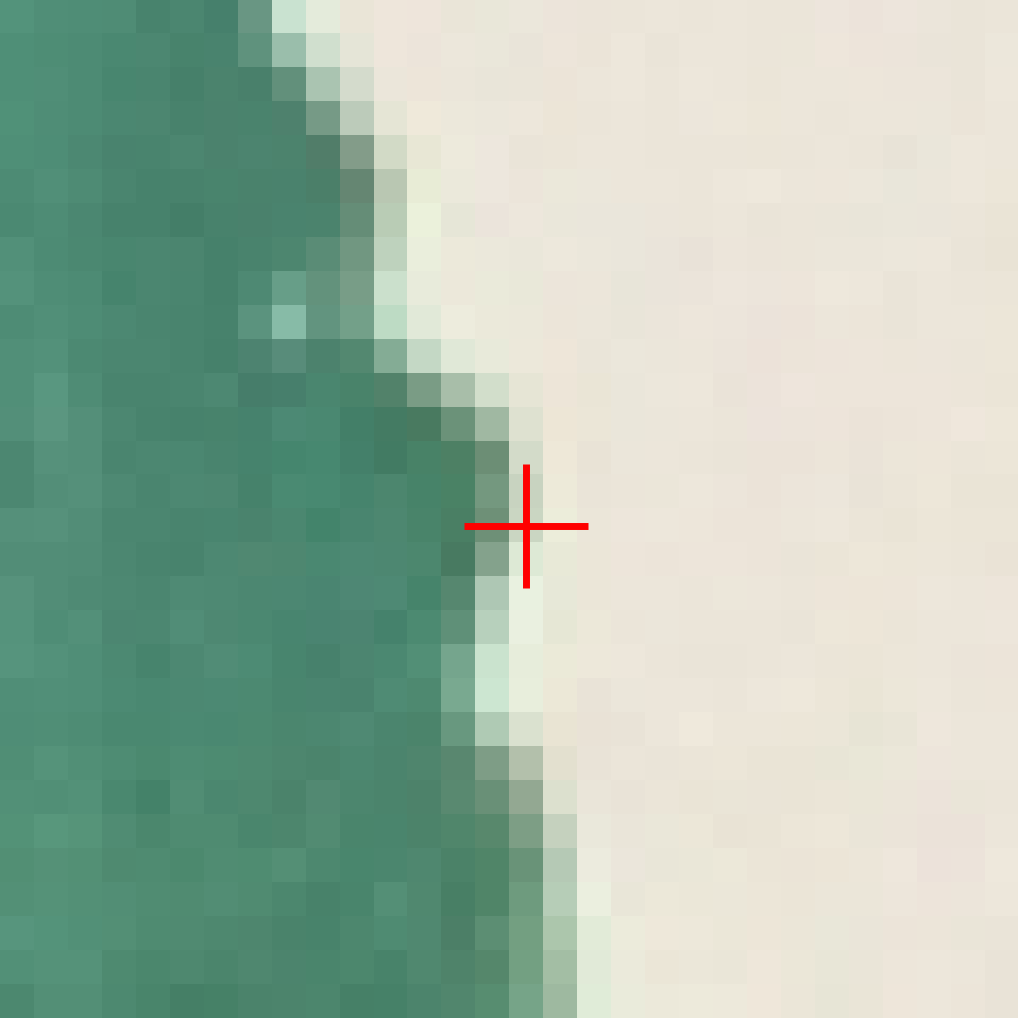}
     \end{minipage} \\
     \begin{minipage}[b]{\mysize}
         \includegraphics[width=\linewidth]{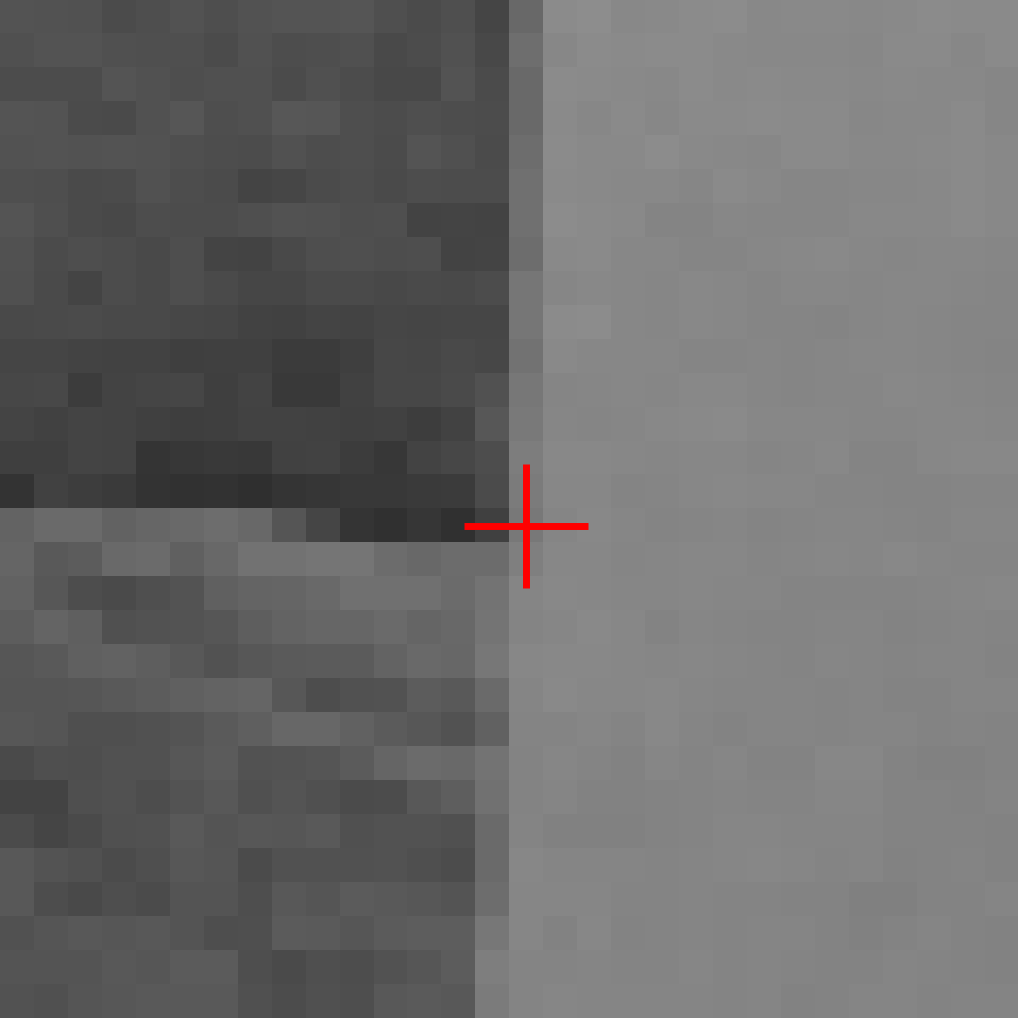}
     \end{minipage}
     \begin{minipage}[b]{\mysize}
         \includegraphics[width=\linewidth]{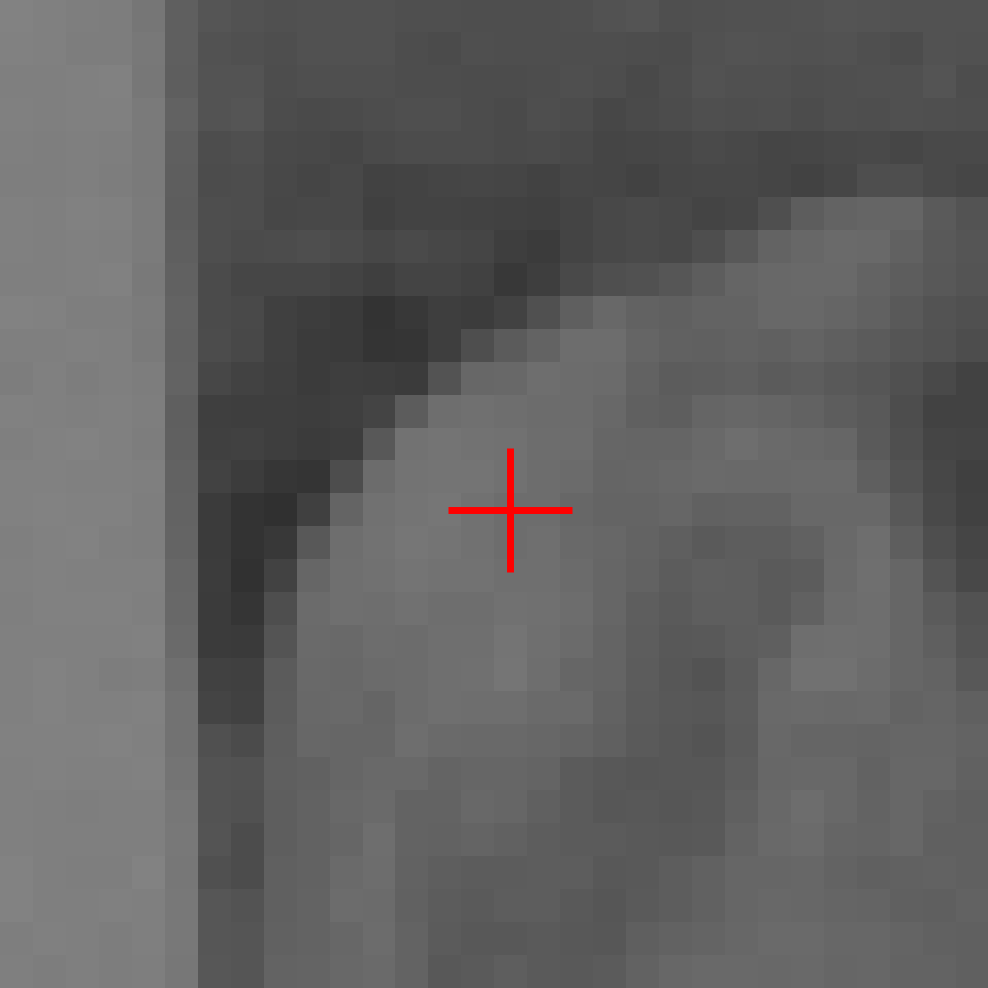}
     \end{minipage}
     \begin{minipage}[b]{\mysize}
         \includegraphics[width=\linewidth]{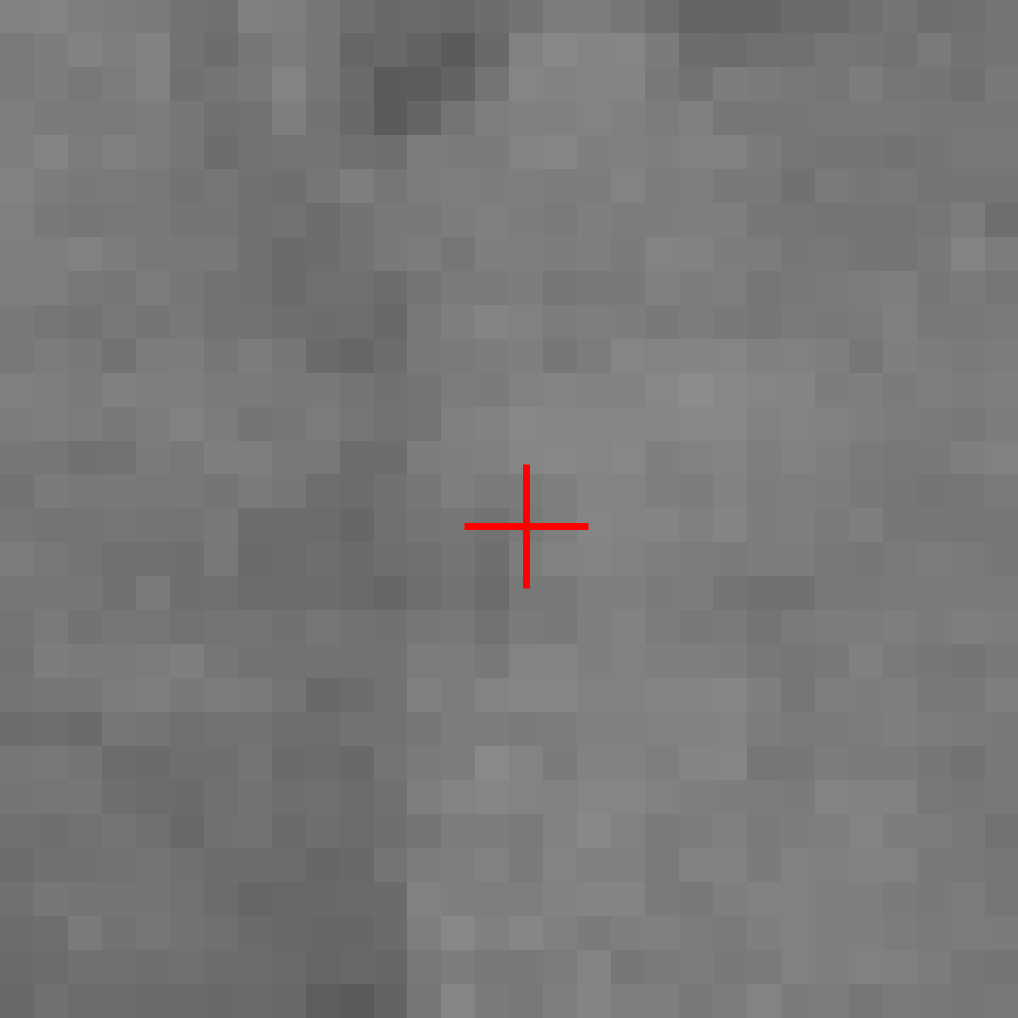}
     \end{minipage}
     \begin{minipage}[b]{\mysize}
         \includegraphics[width=\linewidth]{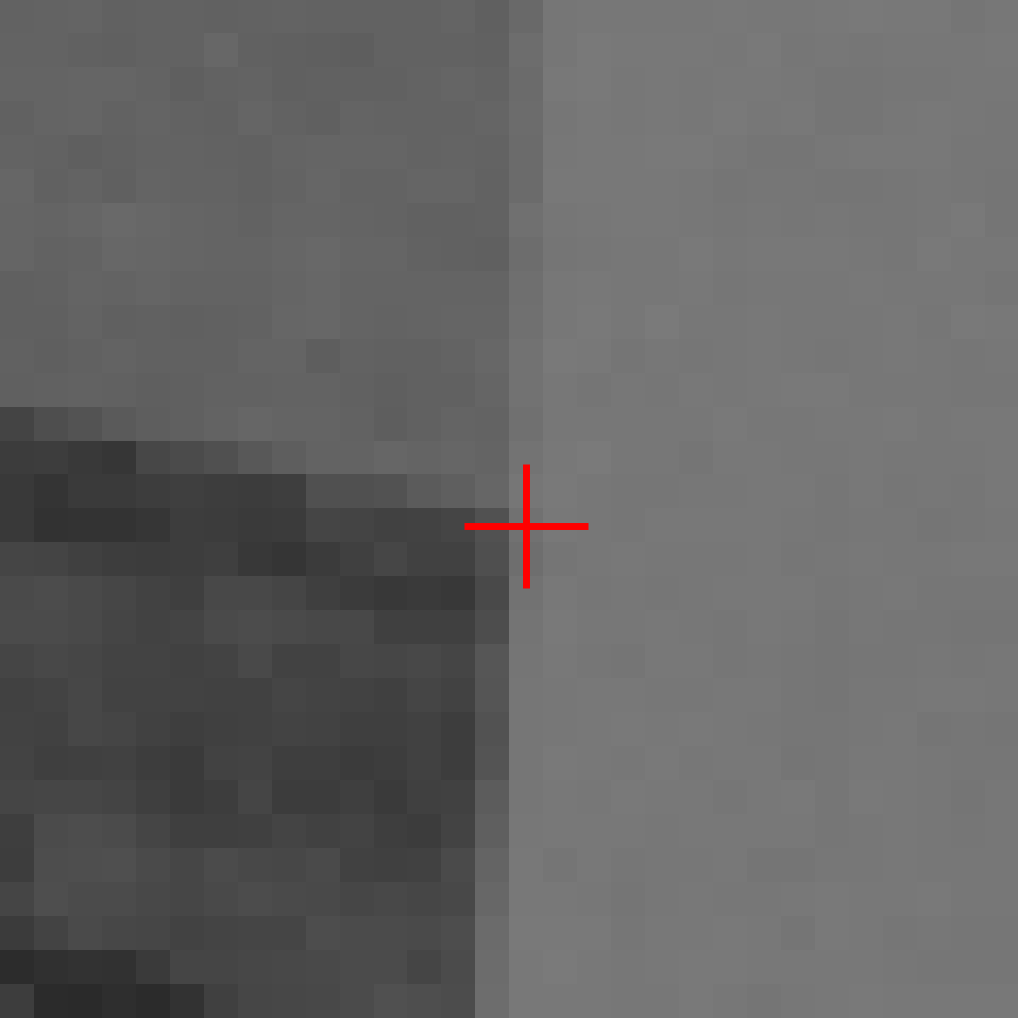}
     \end{minipage}
     \begin{minipage}[b]{\mysize}
         \includegraphics[width=\linewidth]{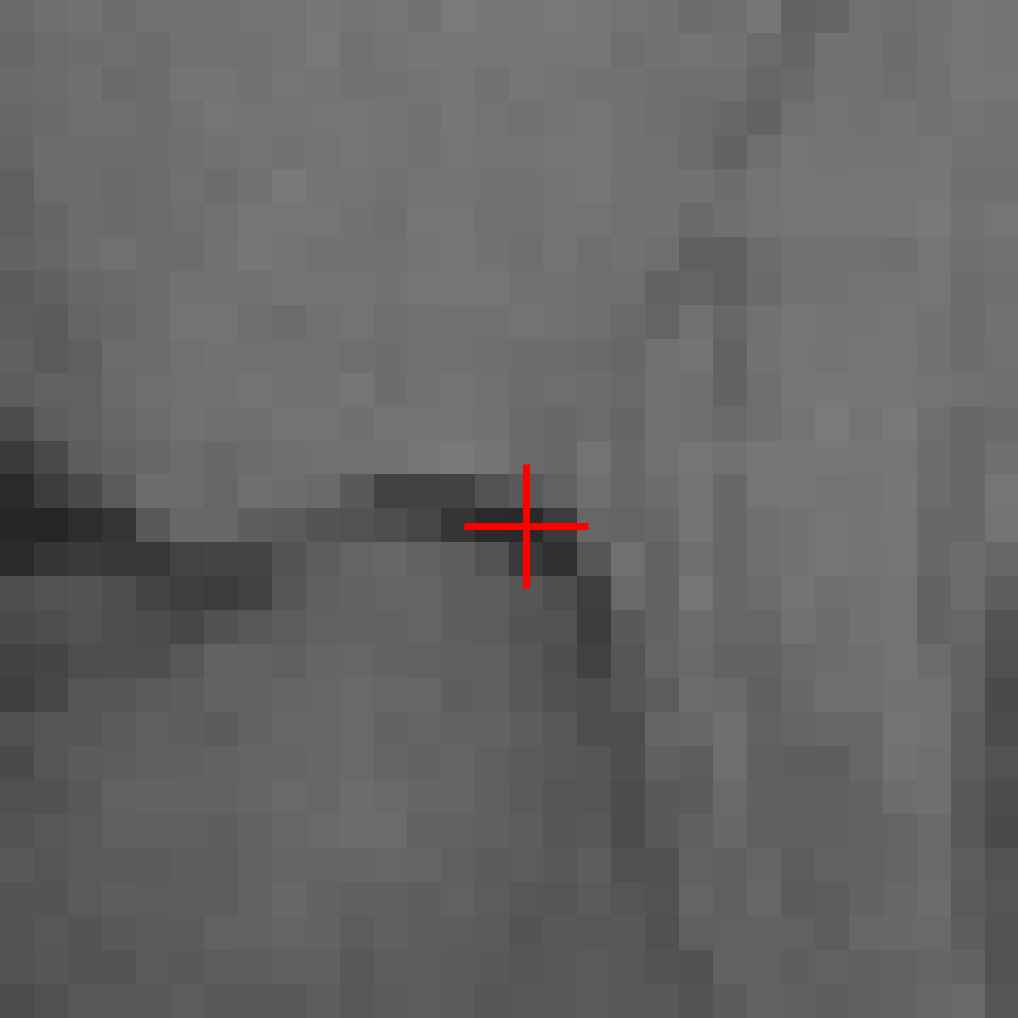}
     \end{minipage}
     \begin{minipage}[b]{\mysize}
         \includegraphics[width=\linewidth]{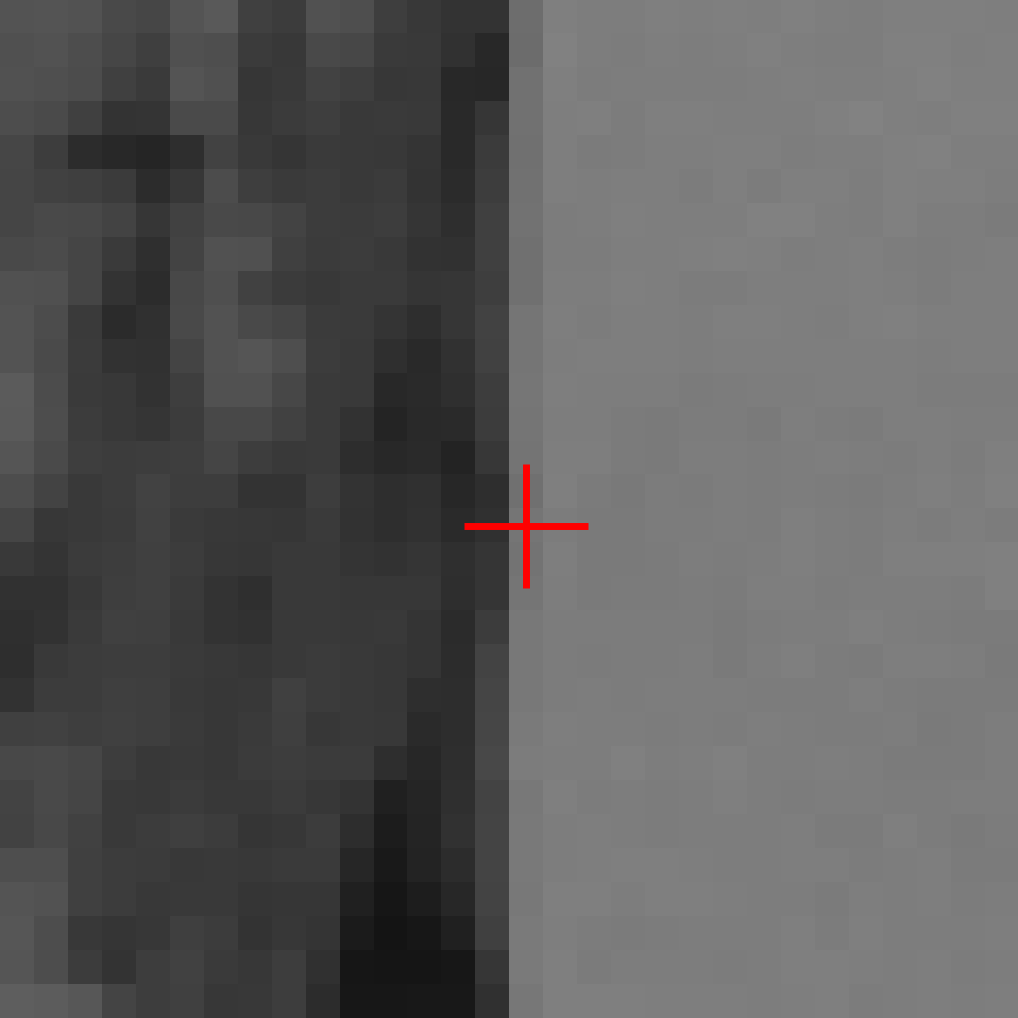}
     \end{minipage} \\
     \begin{minipage}[b]{\mysize}
         \includegraphics[width=\linewidth]{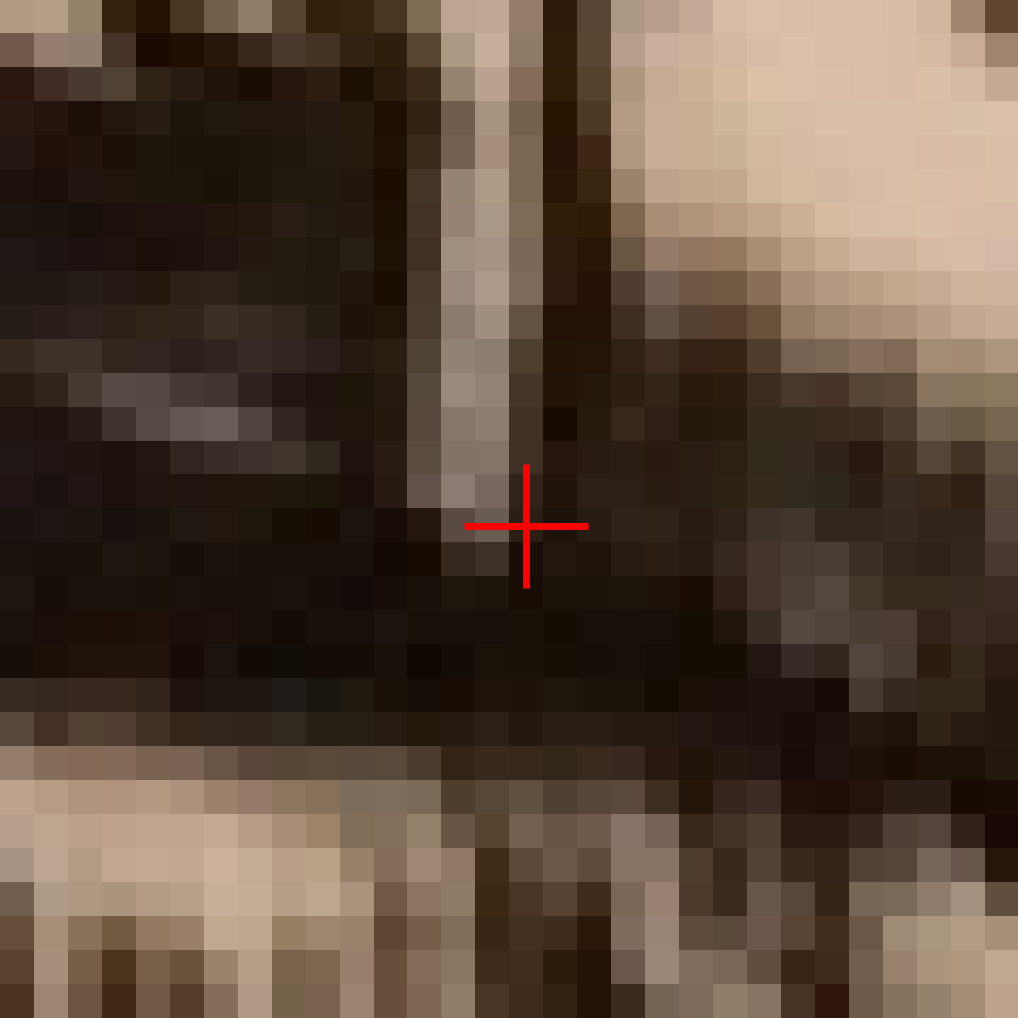}
     \end{minipage}
     \begin{minipage}[b]{\mysize}
         \includegraphics[width=\linewidth]{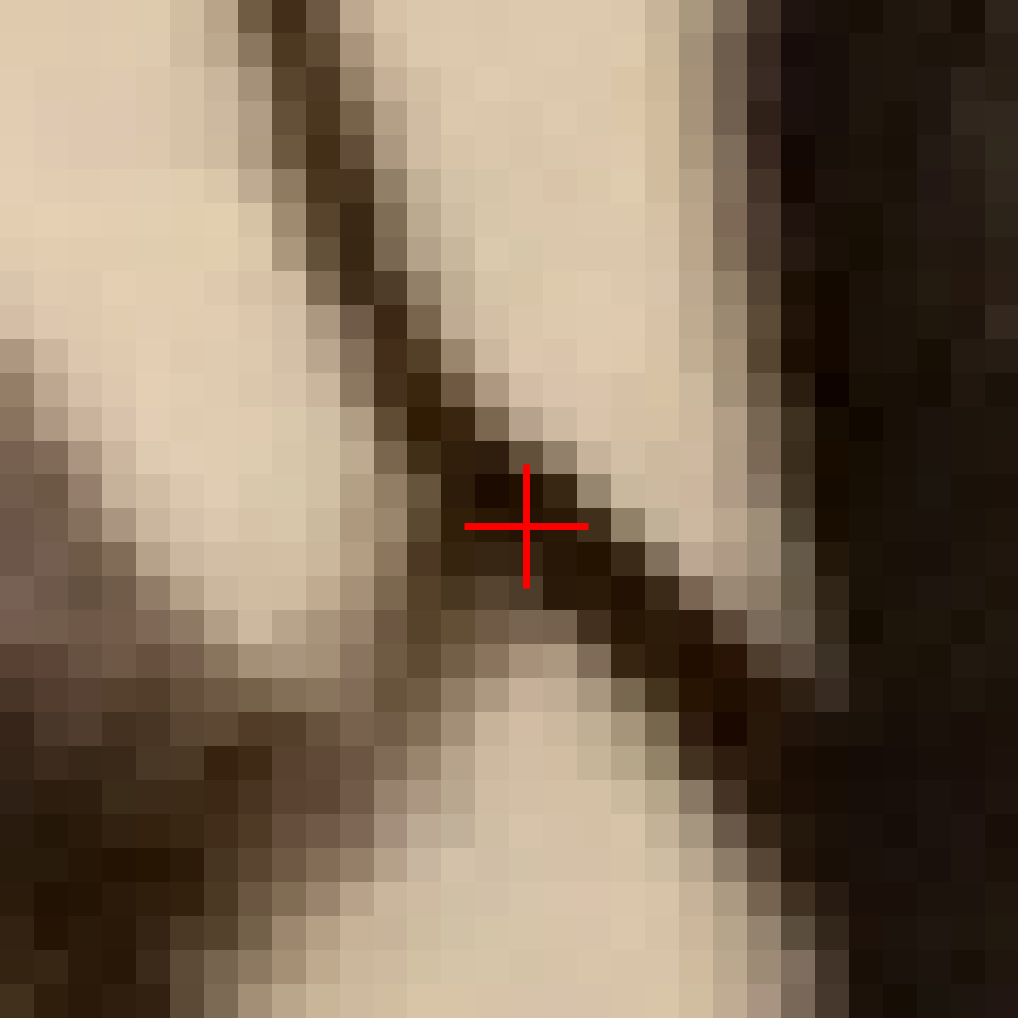}
     \end{minipage}
     \begin{minipage}[b]{\mysize}
         \includegraphics[width=\linewidth]{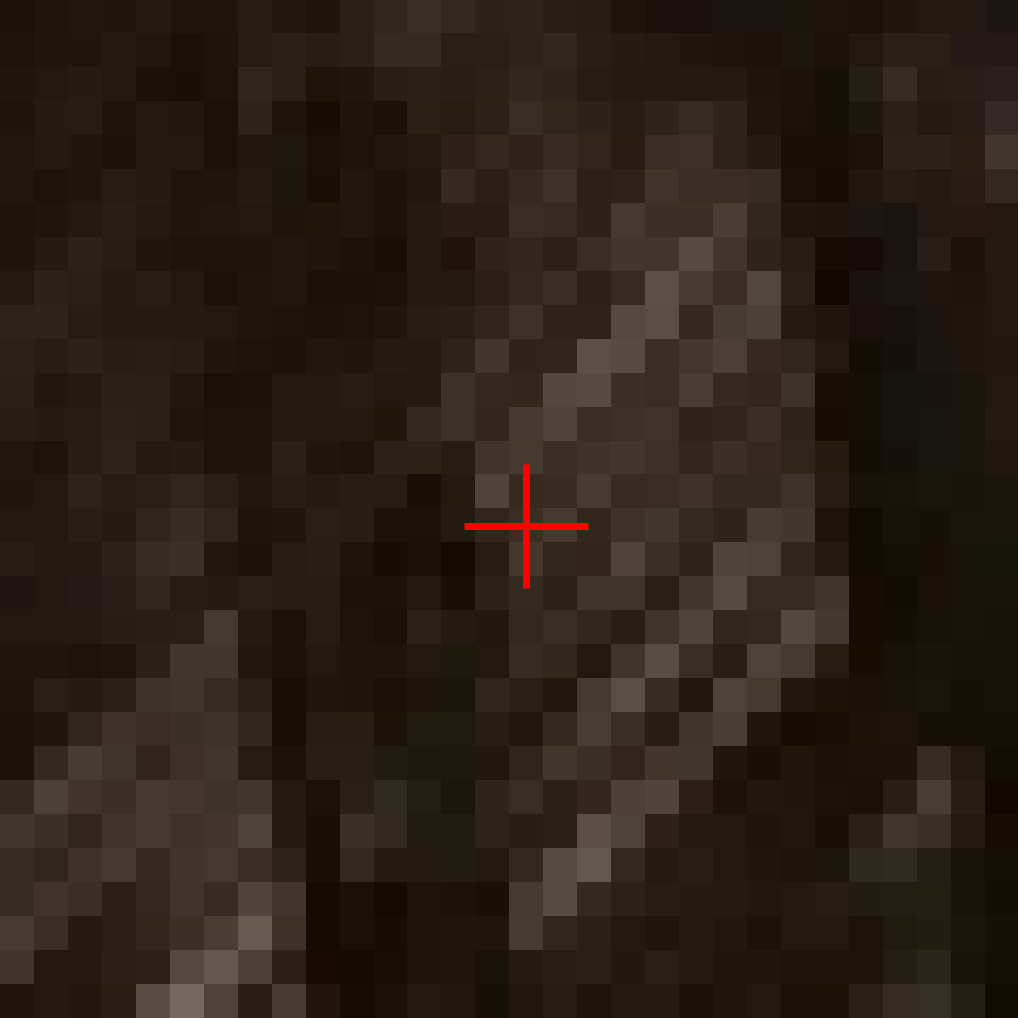}
     \end{minipage}
     \begin{minipage}[b]{\mysize}
         \includegraphics[width=\linewidth]{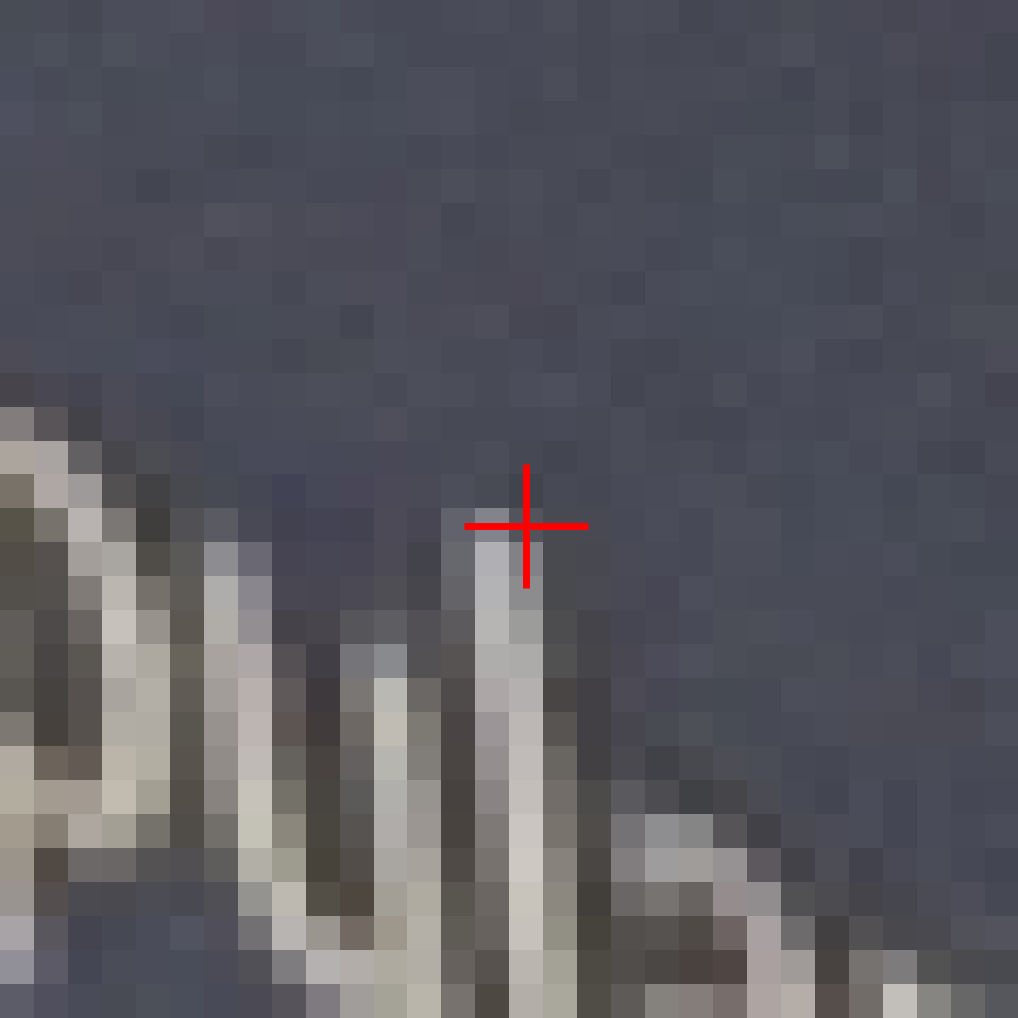}
     \end{minipage}
     \begin{minipage}[b]{\mysize}
         \includegraphics[width=\linewidth]{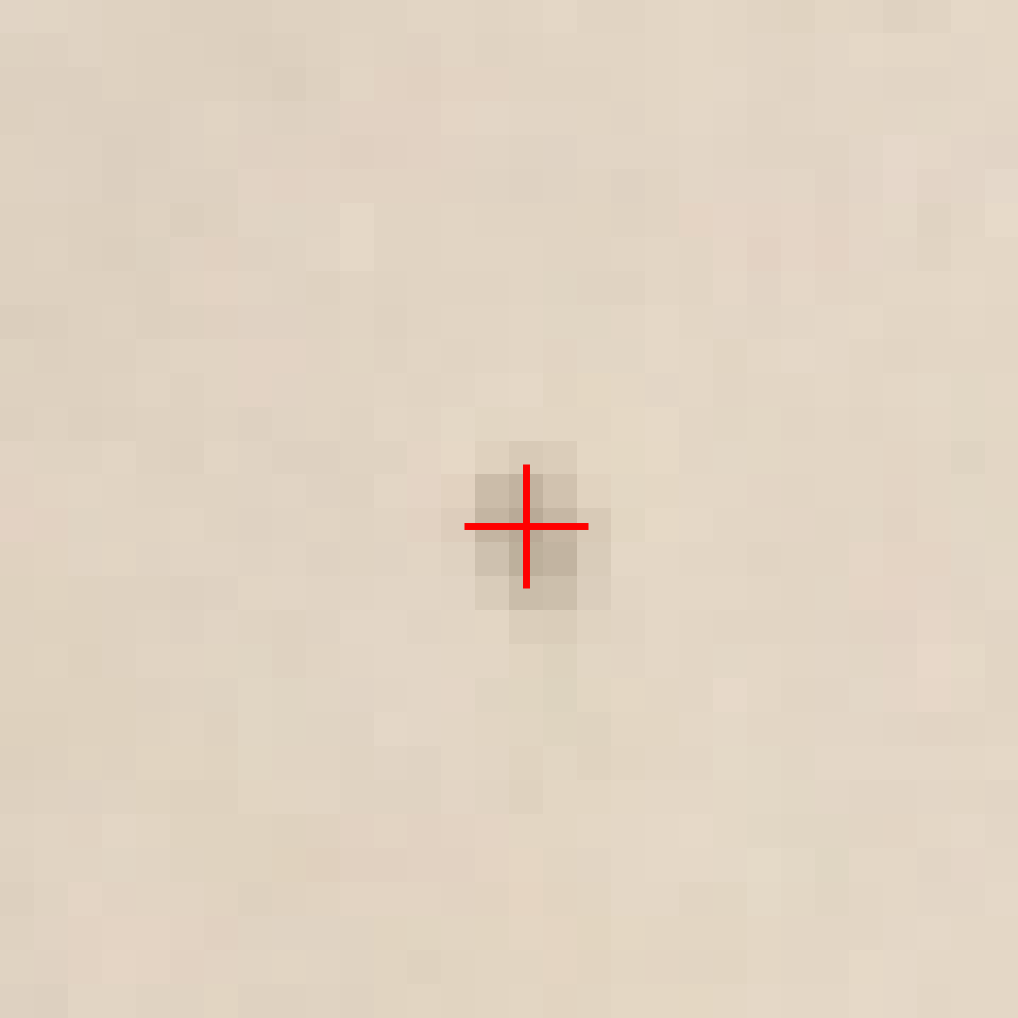}
     \end{minipage}
     \begin{minipage}[b]{\mysize}
         \includegraphics[width=\linewidth]{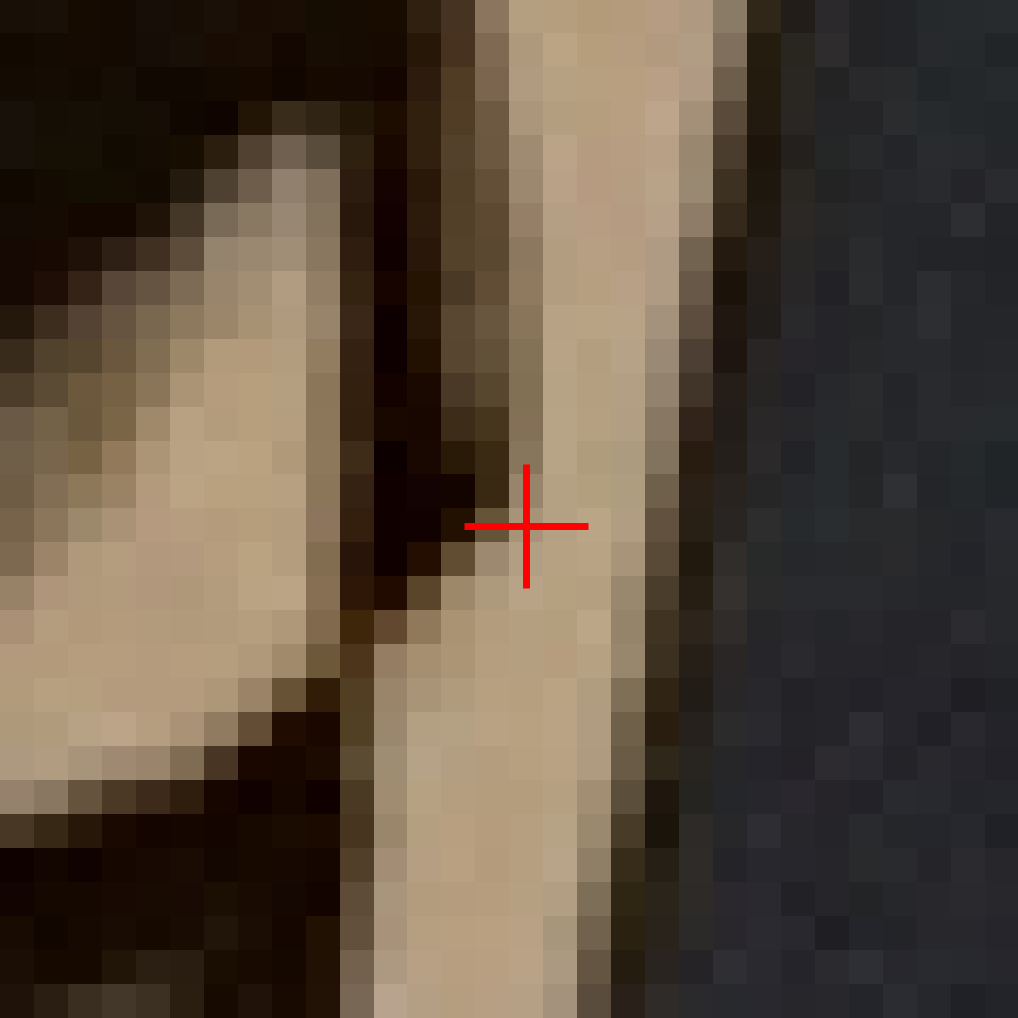}
     \end{minipage} \\
     \begin{minipage}[b]{\mysize}
         \includegraphics[width=\linewidth]{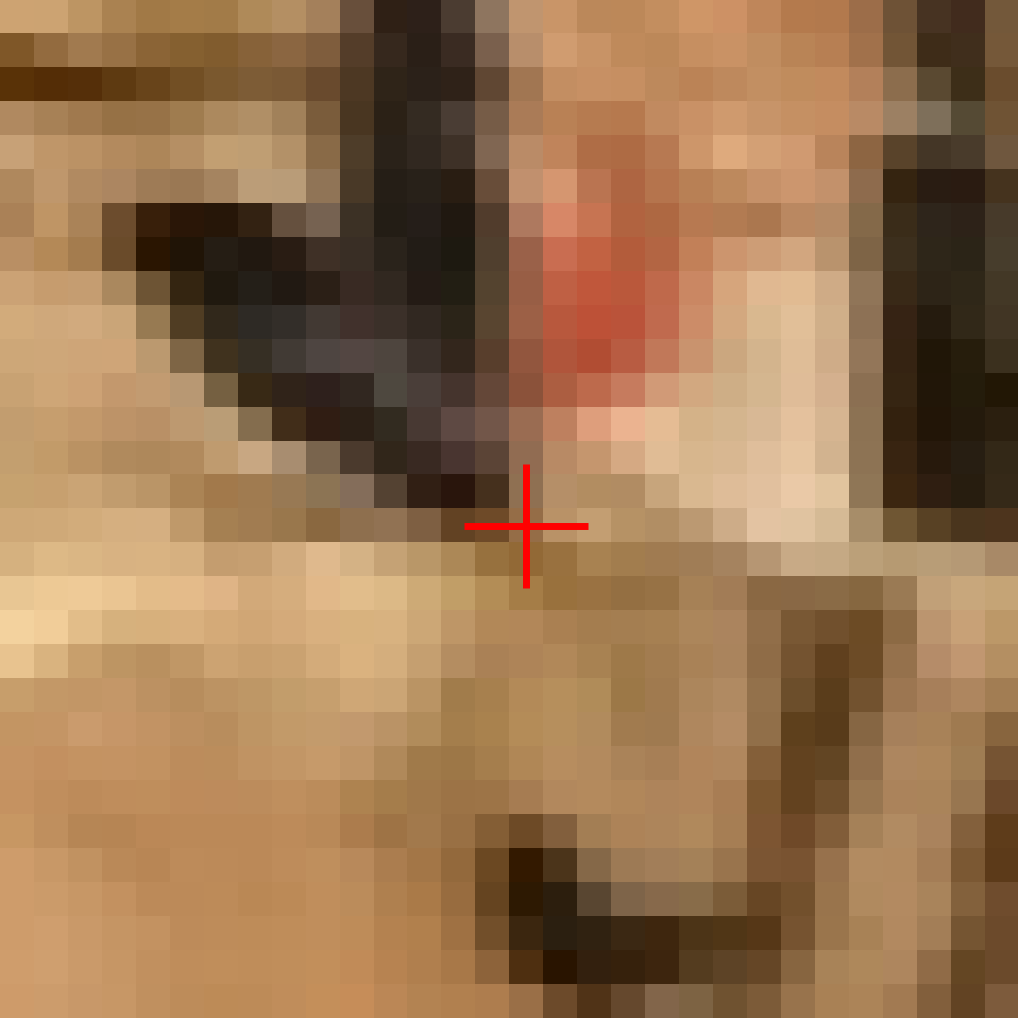}
     \end{minipage}
     \begin{minipage}[b]{\mysize}
         \includegraphics[width=\linewidth]{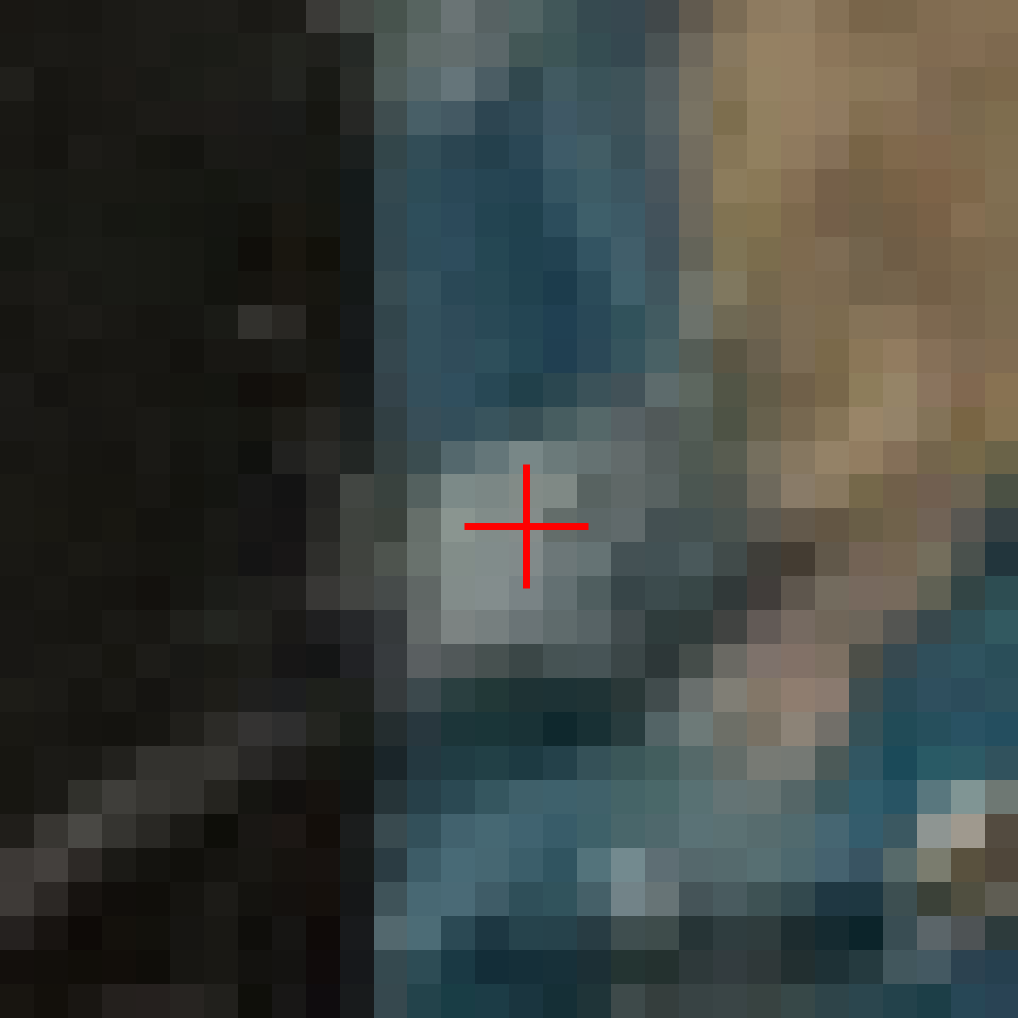}
     \end{minipage}
     \begin{minipage}[b]{\mysize}
         \includegraphics[width=\linewidth]{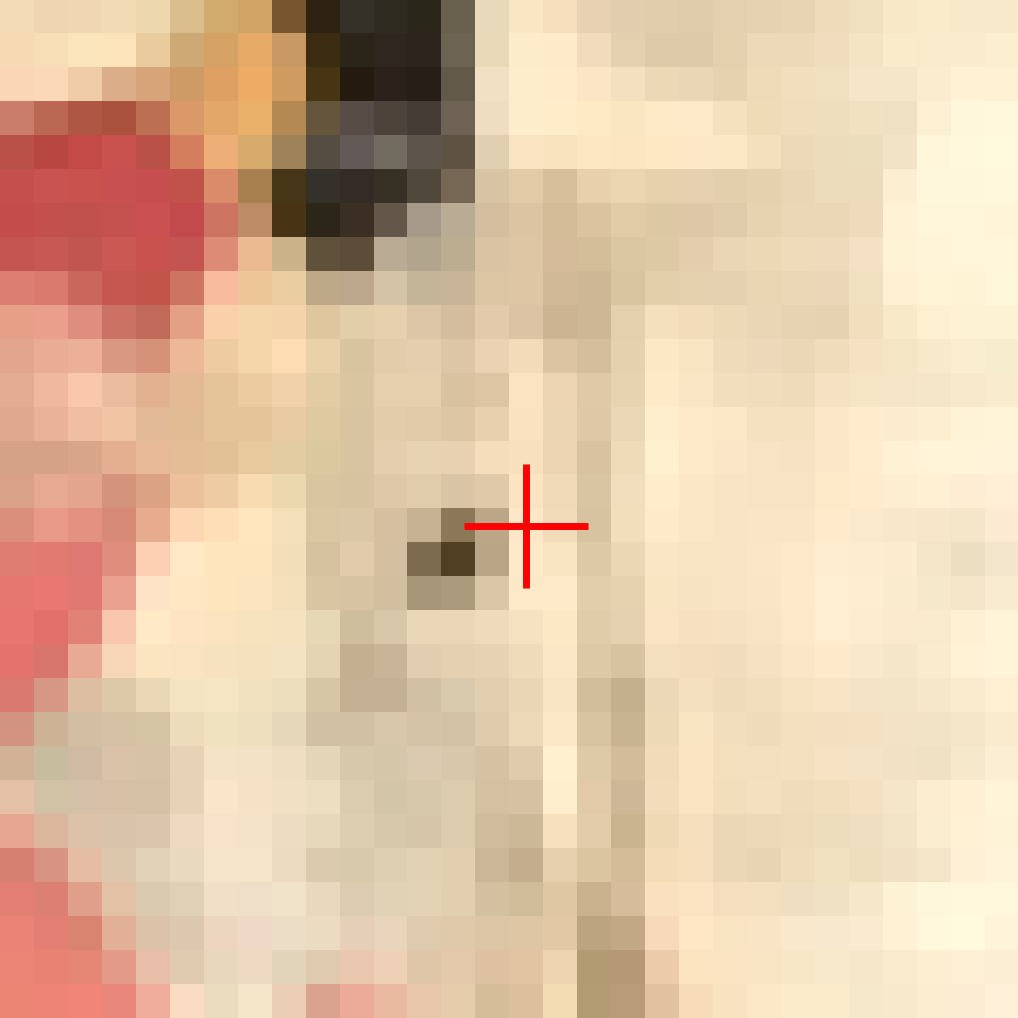}
     \end{minipage}
     \begin{minipage}[b]{\mysize}
         \includegraphics[width=\linewidth]{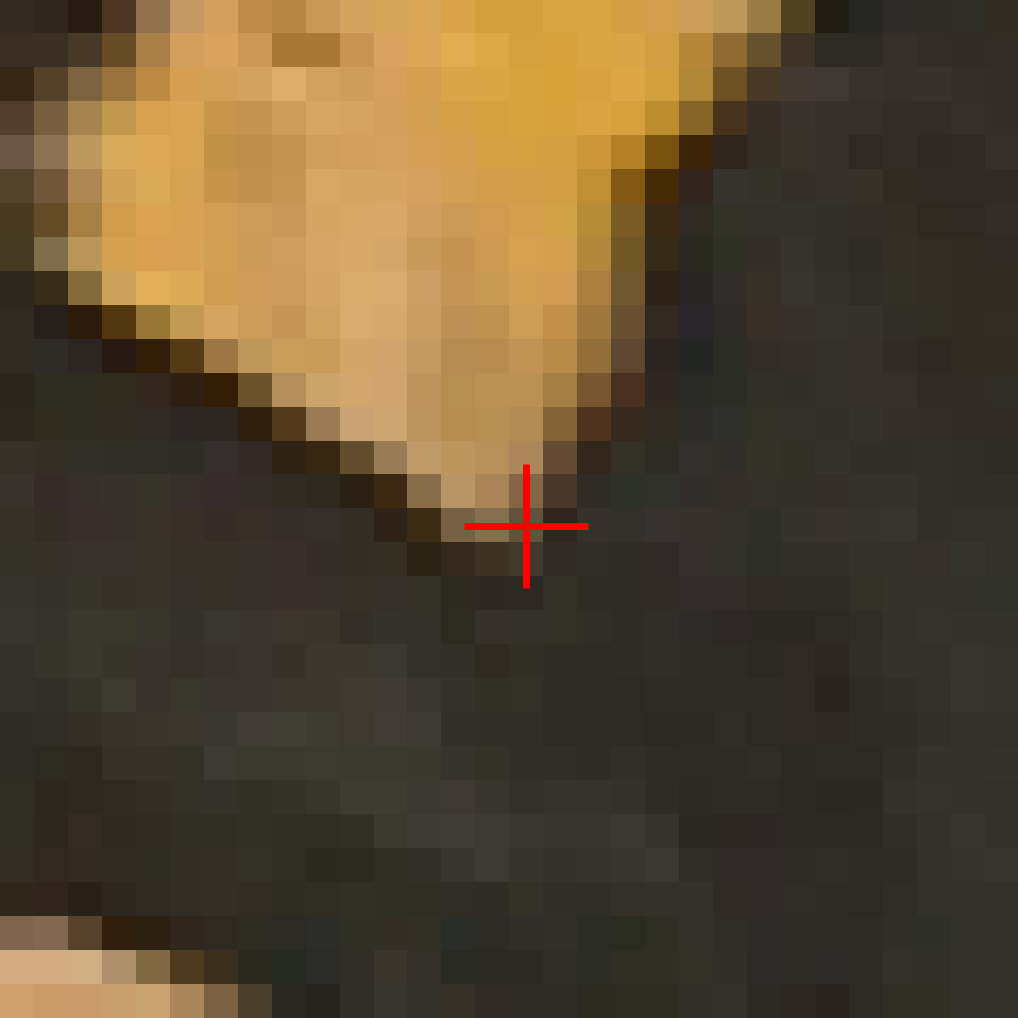}
     \end{minipage}
     \begin{minipage}[b]{\mysize}
         \includegraphics[width=\linewidth]{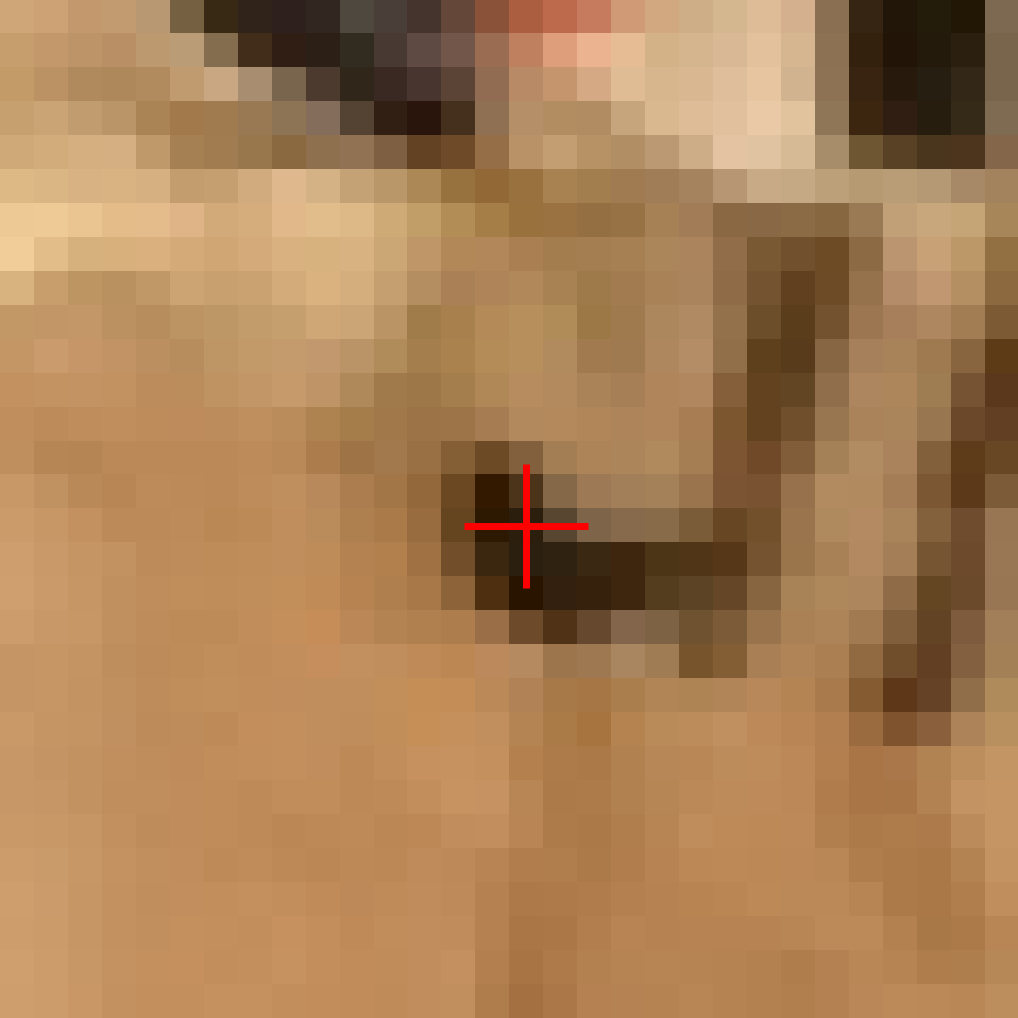}
     \end{minipage}
     \begin{minipage}[b]{\mysize}
         \includegraphics[width=\linewidth]{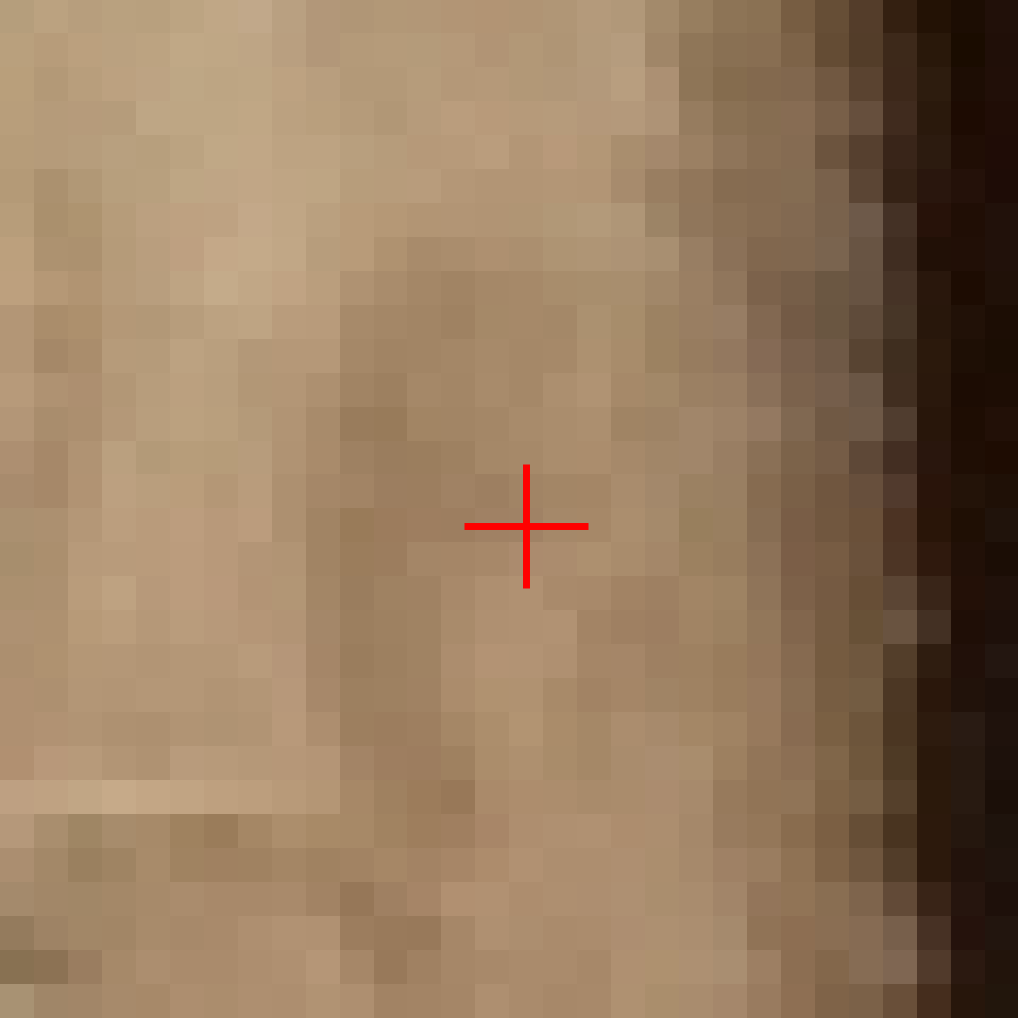}
     \end{minipage} \\
     \begin{minipage}[b]{\mysize}
         \centering
         \includegraphics[width=\linewidth]{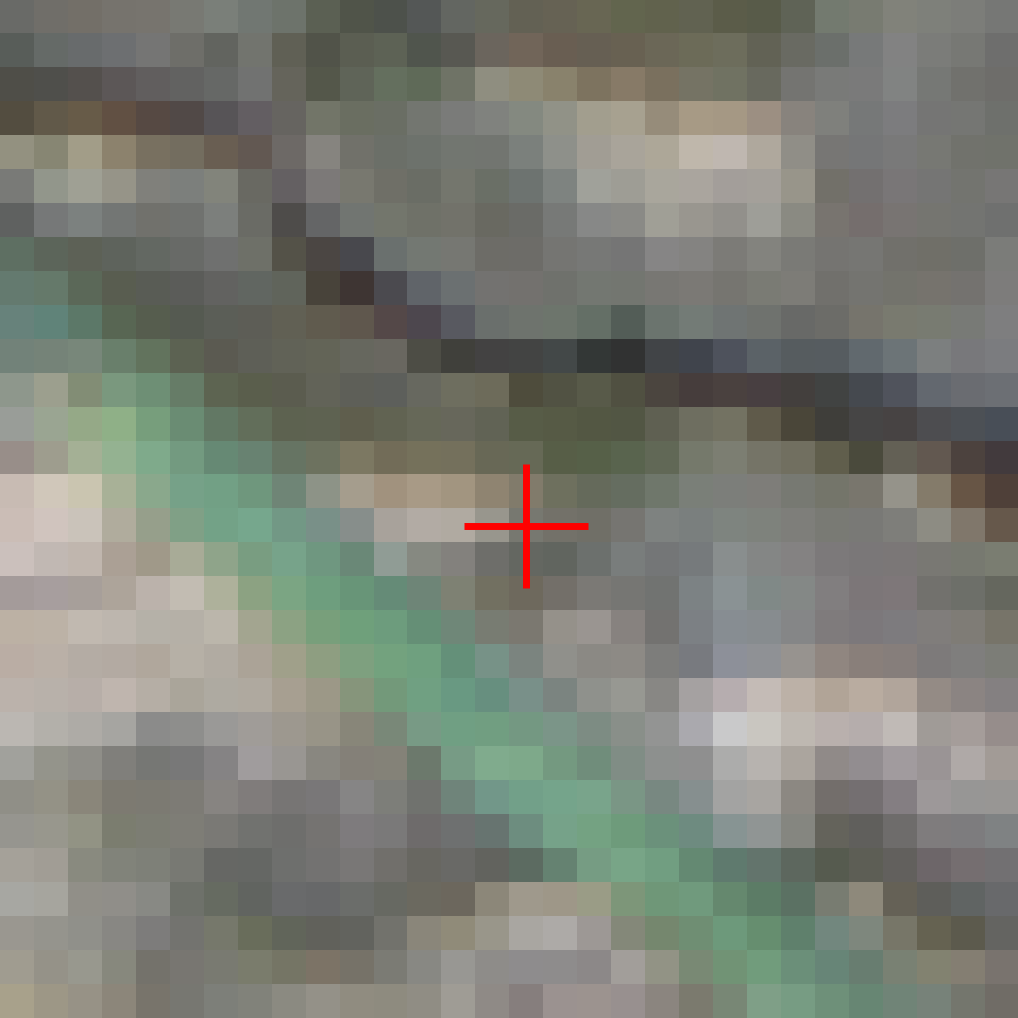} \\
         \tiny Harris
     \end{minipage}
     \begin{minipage}[b]{\mysize}
         \centering
         \includegraphics[width=\linewidth]{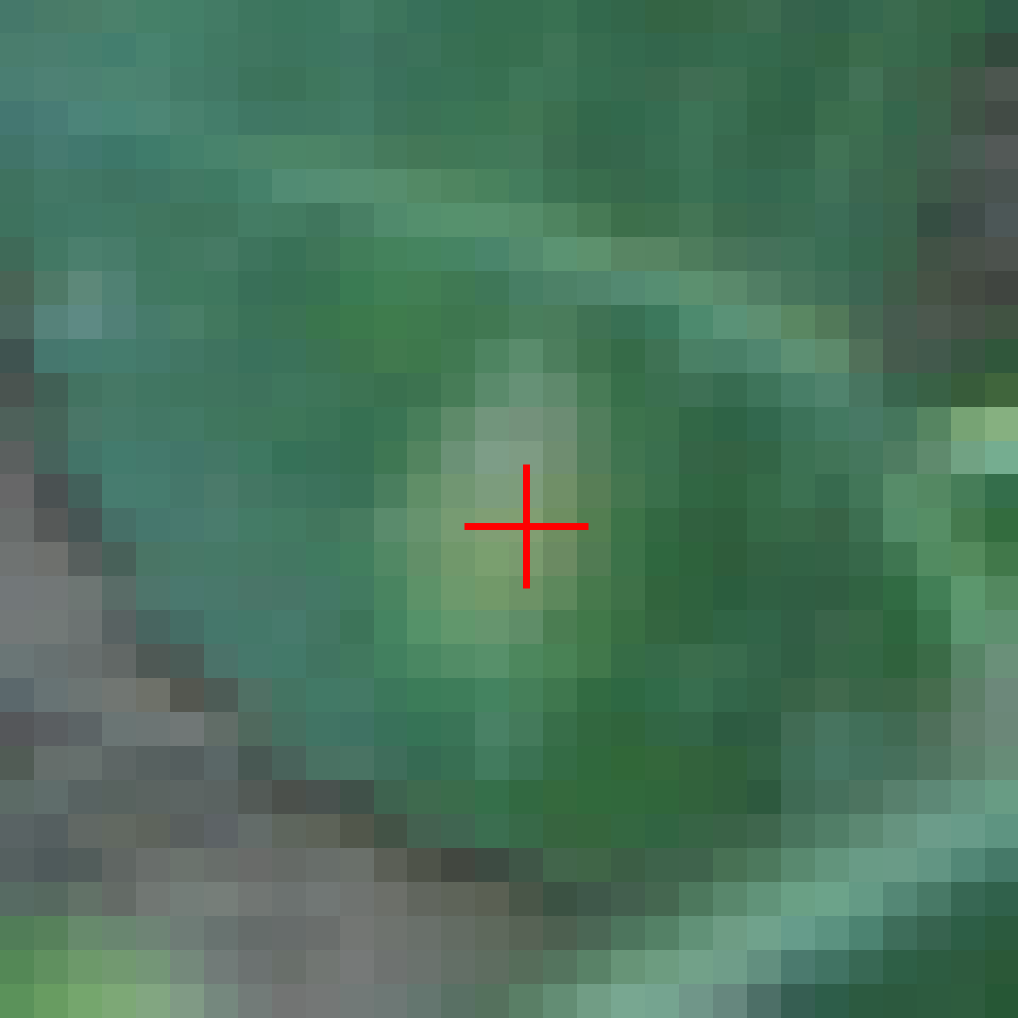} \\
         \tiny DoG
     \end{minipage}
     \begin{minipage}[b]{\mysize}
         \centering
         \includegraphics[width=\linewidth]{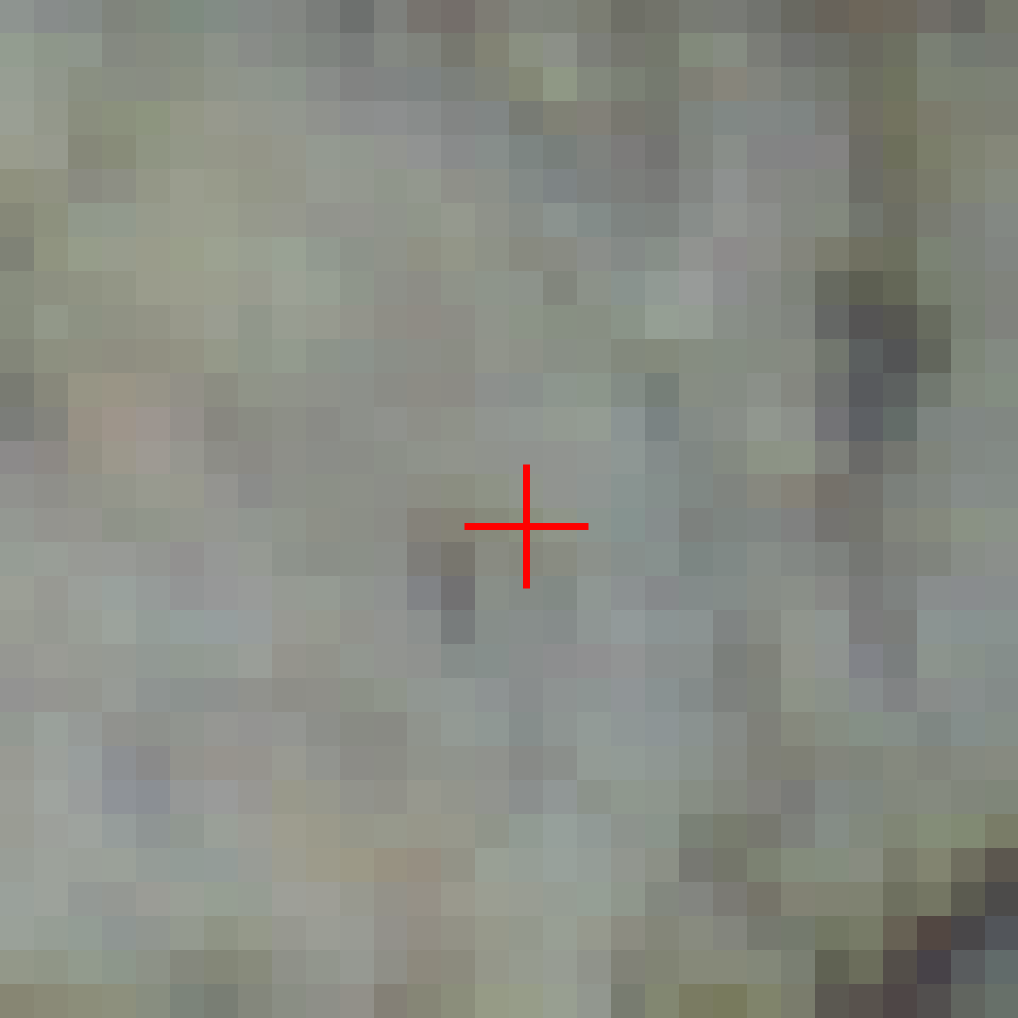} \\
         \tiny DISK
     \end{minipage}
     \begin{minipage}[b]{\mysize}
         \centering
         \includegraphics[width=\linewidth]{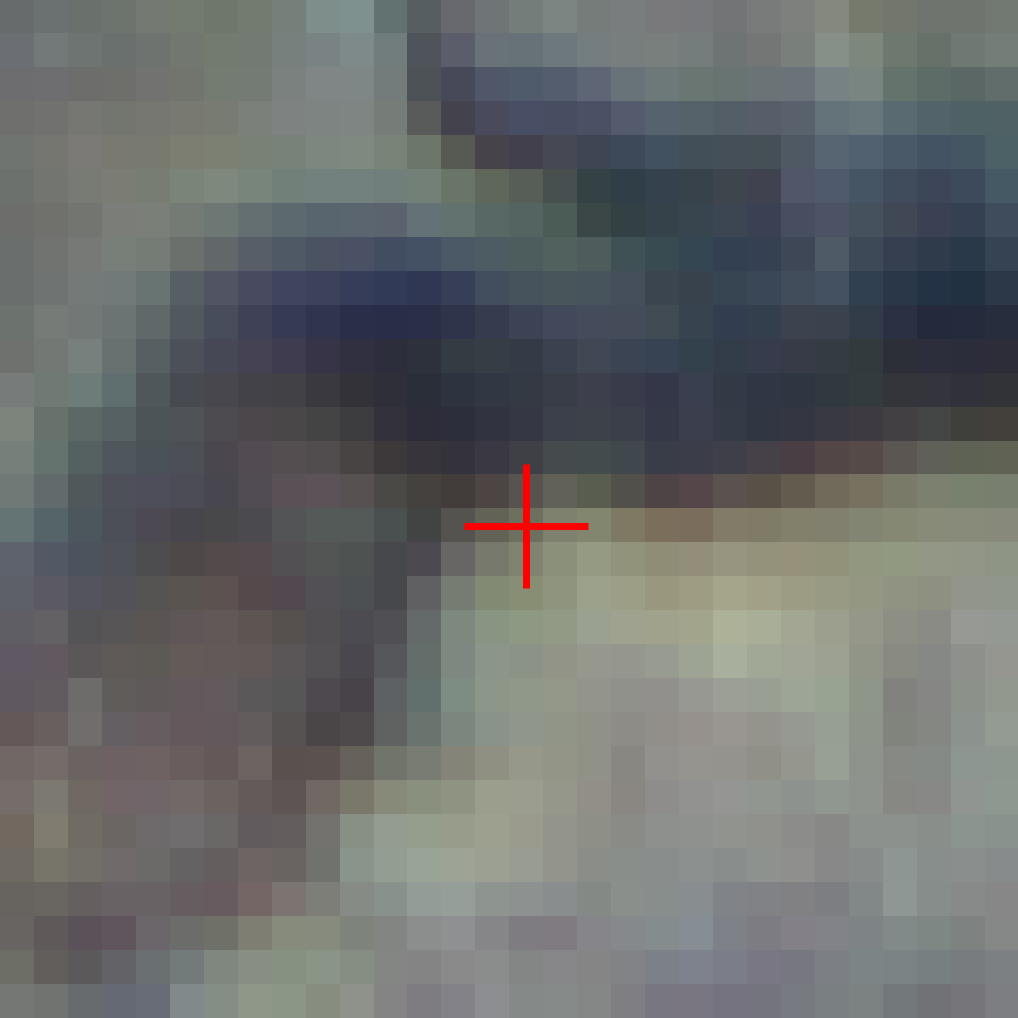} \\
         \tiny SuperP.
     \end{minipage}
     \begin{minipage}[b]{\mysize}
         \centering
         \includegraphics[width=\linewidth]{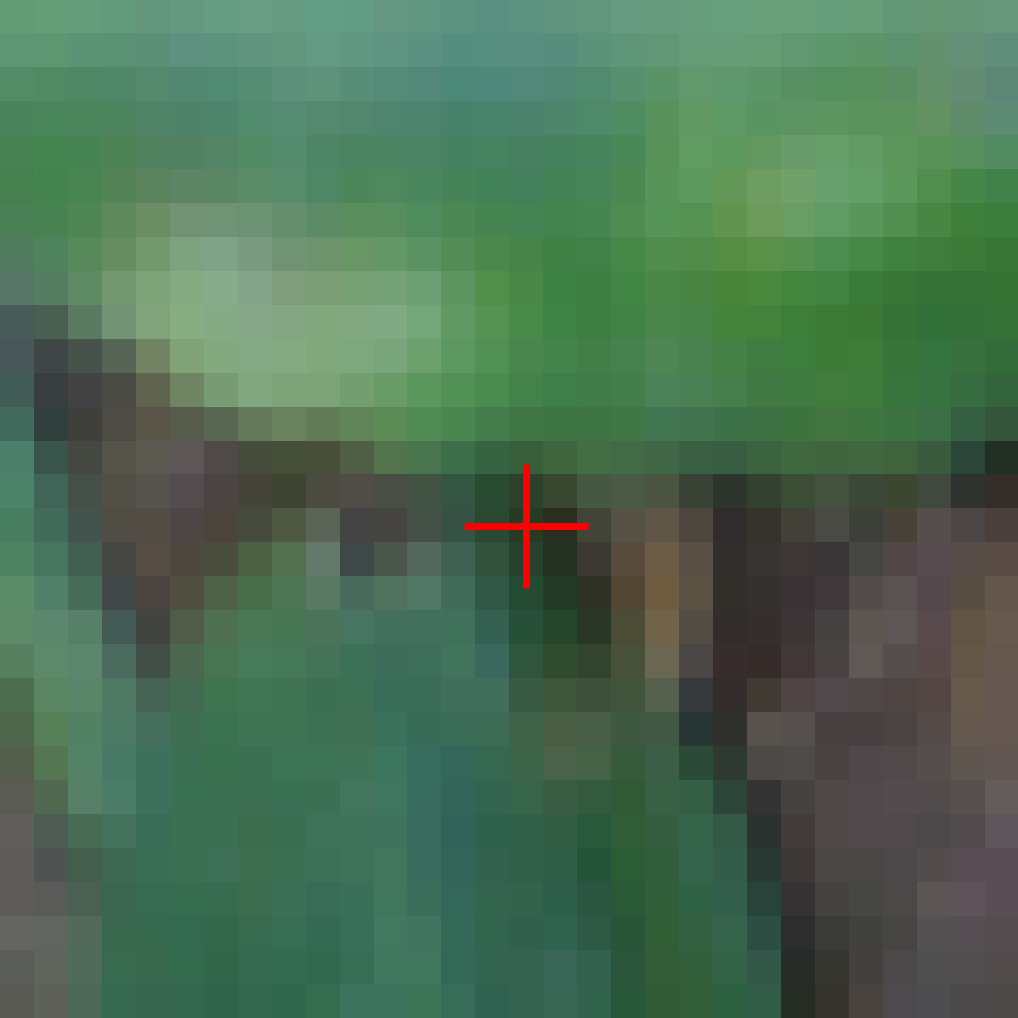} \\
         \tiny R2D2
     \end{minipage}
     \begin{minipage}[b]{\mysize}
         \centering
         \includegraphics[width=\linewidth]{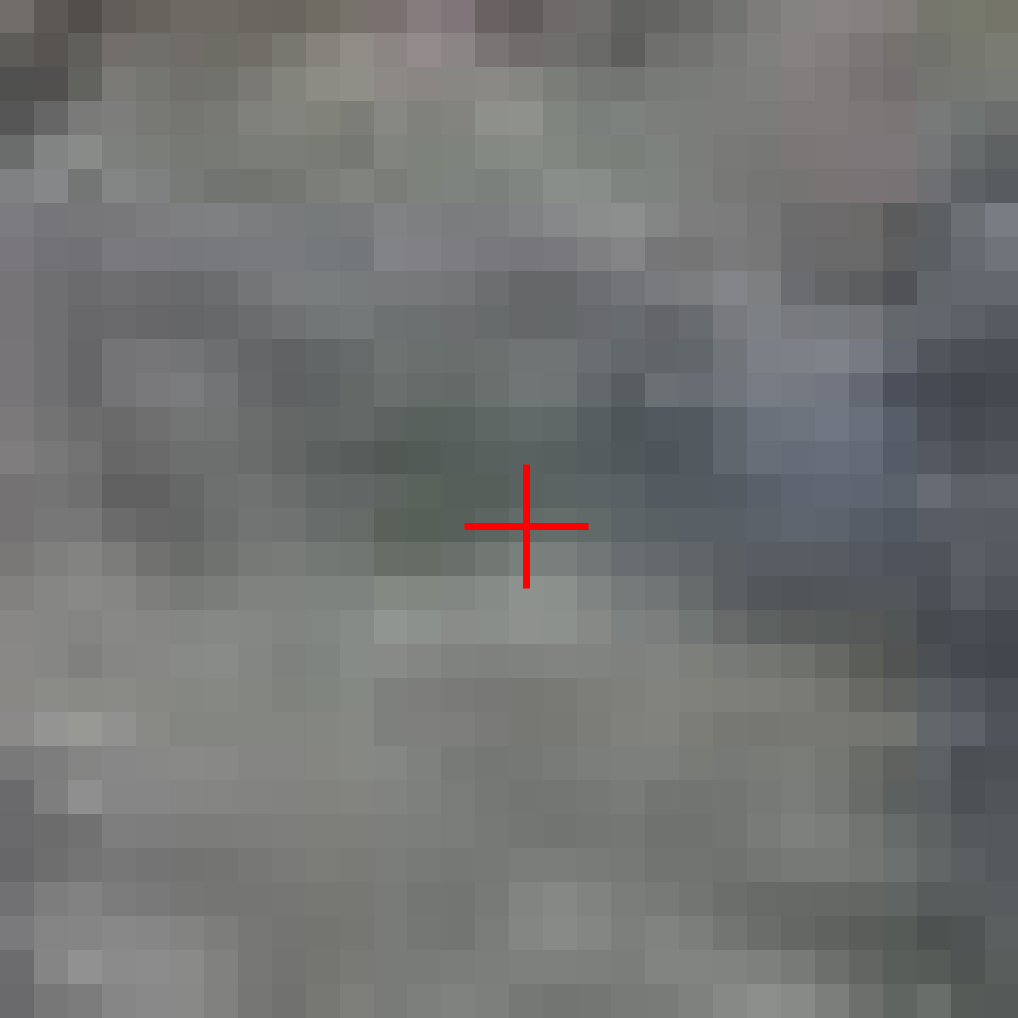} \\
         \tiny MD-Net
     \end{minipage} \\
    \captionof{figure}{Qualitative comparison between the best clusters found by different methods in the first 5 scenes of HPatches.
    For each scene (one per row), we select the lowest \emph{deviation} cluster (best localization accuracy) 
    among all the ones with \emph{robustness} = \num{21} 
    (which represents keypoints that have been detected in all warps).
    }
    \label{fig:best_keypoints}
\end{minipage}
\end{figure}

\section{Method}
The purpose or our method is twofold:
it refines the positions of a given keypoint set,
additionally dropping low quality keypoints and adding new stable ones,
while at the same time characterizing each resulting keypoint with two interpretable scores.
These scores directly relate to two orthogonal keypoint properties: 
robustness to viewpoint changes and localization accuracy.
As depicted in Figure~\ref{fig:pipeline},
our pipeline consists of several stages,
which we detail in the following subsections.

\subsection{Multiple Image Warping}
The goal of the first stage of our pipeline is to simulate real-world viewpoint changes.
To this end, we generate multiple altered versions of the original image
by applying a series of affine warpings, each sampled from a fixed set of possible transformations.
In order to avoid uneven keypoint densities in the next step,
we only use transformations that distort every part of the image uniformly.

\subsection{Keypoint Detection \& Reprojection}
We run the chosen keypoint detector on all the warped images
and pick the best \textit{n} keypoints from each, according to the scores given by the detector.
Each set of detected keypoints is then re-projected into the input image reference frame
through the inverse of the transformation originally used to generate the warped image.
To compensate for very close keypoints,
which may happen as a consequence of warpings that increase the image area,
we apply a \gls{nms} to the reprojected keypoints from each warped image separately.
Keypoints retained after \gls{nms} are then aggregated into a single set, 
regardless of the warped image they have been extracted from.

\subsection{Density Estimation}
The proposed approach characterizes keypoints based 
on robustness against viewpoint changes and localization accuracy,
which we describe by metrics derived from a \gls{gmm}.
To ensure a proper fit, the parameters and initialization of the \gls{gmm},
in particular the number of components and their respective means,
have to be chosen in an informed manner.
In this section,
we briefly describe how we utilize classical \gls{kde} to initialize the down-stream fitting task.

Let \( (x_i)_{i=1}^N \subset \R^2 \) denote the collection of \( N \) re-projected keypoints extracted from the augmentation pipeline.
From this collection, we construct a \gls{kde} \( f_{\text{KDE}} : \R^2 \to \R_+ \) as
\begin{equation}
    f_{\text{KDE}}(x) = (Nh^2)^{-1}\sum_{i=1}^N \phi\biggl( \frac{x_i - x}{h} \biggr).
\end{equation}
We utilize the symmetric Gaussian kernel \( \phi : \R^2 \to \R_+ : \)
\begin{equation}
    \phi(x) = (2\pi h)^{-1}\exp\bigl( -\norm{x}^2 / 2 \bigr).
\end{equation} 
In the definitions above, \( h \in \R_+ \) is the \gls{kde} bandwidth in pixels, the choice of which determines the fidelity of \( f_{\text{KDE}} \).
We found that the choice of \( h \) is not crucial (within a reasonable range) and barely influences the subsequent \gls{gmm} fitting routine.

For the \gls{gmm} fitting, we then utilize the set of maximizers \( \argmax f_{\text{KDE}} \) as the initialization for the component means.
A classical algorithm to find \( \argmax f_{\text{KDE}} \) is the mean shift algorithm~\cite{meanshift}.
However, it is computationally demanding and there is no strict requirement for the resulting points to be exact maximizers,
as they only serve as an initialization to the subsequent fitting algorithm.
Thus, we resort to evaluating \( f_{\text{KDE}} \) on the regular Cartesian grid \( \Omega = \{ 1, \dotsc, W \} \times \{ 1, \dotsc, H \} \) of pixel centers for an image with dimensions \( W \times H \) and find maxima on this grid in a local neighborhood.
In particular, we find the points \( \tilde{\mathcal{X}}_{\text{max}} = \{ x \in \Omega : f_{\text{KDE}}(x) > \zeta, f_{\text{KDE}}(y) < f_{\text{KDE}}(x) \,\forall y \in \omega_r(x) \} \), where \( \zeta \in \R_+ \) is a threshold to avoid low-density regions and \( \omega_r(x) = \{ y \in \Omega : \norm{y - x}_\infty \leq r \} \) is a window of radius \( r \in \mathbb{N} \) centered around \( x \).
To have an upper bound on the components in the subsequent \gls{gmm} fitting routine, we take the \( 2 N_{\text{kpts}} \) points that score best under \( f \): \( \mathcal{X}_{\text{max}} = \{ x_i \in \tilde{\mathcal{X}}_{\text{max}} : i \in \mathcal{I} \} \), where \( \mathcal{I} \) are the \( 2N_{\text{kpts}} \) indices into \( \tilde{\mathcal{X}}_{\text{max}} \) such that \( f_{\text{KDE}}(x_i) > f_{\text{KDE}}(x_j) \), for all \( i \in \mathcal{I} \) and all \( j \notin \mathcal{I} \).

\noindent
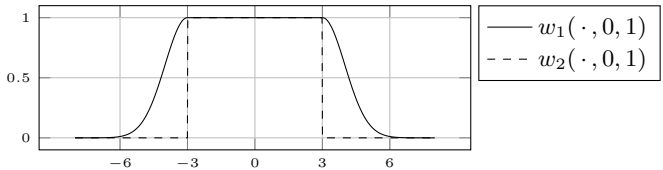
\begin{figure}[t]
\centering
\begin{tikzpicture}[
    declare function={%
        w1(\x,\m,\s)=(abs(\x - \m) < 3 * \s) + (abs(\x - \m) > 3 * \s) * exp(-(abs(\x - \m) - 3 * \s^2)^2 / (2 * \s^2));
        w2(\x,\m,\s)=(abs(\x - \m) < 3 * \s);
    },
]
    \begin{axis}[
        domain=-8:8,
        samples=1000,
        height=3.5cm,
        width=0.6\textwidth,
        grid,
        xtick={-6,-3,0,3,6},
        legend style={at={(1.44,0.74)},anchor=east},
    ]
        \addplot[black] {w1(x,0,1)};
        \addplot[black, dashed] {w2(x,0,1)};
        \legend{\( w_1(\,\cdot\,, 0, 1) \)\\ \( w_2(\,\cdot\,, 0, 1) \)\\}
    \end{axis}
\end{tikzpicture}
\captionof{figure}{Outlier weighting functions}%
\label{fig:weighting}
\end{figure}

\subsection{Gaussian Mixture Model}
To refine keypoint locations and assign scores, we leverage a \gls{gmm}.
We chose this model as its parameters inherently allow us to refine and score keypoints in an \emph{interpretable} fashion.
In particular, refined keypoint positions are given by the means of the components of the \gls{gmm},
and for a specific component, 
\emph{robustness} is measured via its weight and the \emph{deviation} via its variance.
As the negative influence of outliers needs to be considered in our setting,
we propose a custom robust fitting algorithm, which we elaborate on in this section.

A \( K \)-component \gls{gmm}
\begin{equation}
    \sum_{k=1}^K \alpha_k G_{\mu_k, \Sigma_k}
\end{equation}
is a convex combination of \( K \) Gaussians
\begin{equation}
G_{\mu, \Sigma} : \R^d \to \R_+ : x \mapsto \frac{\exp\bigl( -\norm{x - \mu}^2_{\Sigma^{-1}} \bigr)}{\sqrt{\det (2\pi \Sigma)}}
\end{equation}
parameterized by their means \( \mu \in \R^d \) and covariance matrices \( \Sigma \in \mathbb{S}^d_+ \).
For proper normalization, the weight vector \( \alpha = (\alpha_1,\dotsc,\alpha_K)^\top \) must satisfy \( \alpha \geq 0 \), \( \langle \alpha, \mathds{1}_d \rangle = 1 \).
In our case, since we are fitting a \gls{gmm} on the image domain \( d = 2 \), we restrict ourselves to isotropic covariance matrices of the form \( \Sigma_k = \sigma_k^2 \mathrm{Id}_2 \).
We discuss the implications of this choice later and, with slight abuse of notation, denote with \( G_{\mu,\sigma} \) a Gaussian with mean \( \mu \) and covariance matrix \( \sigma^2\mathrm{Id} \).

The most widely used estimator for the parameters \( (\alpha_k, \mu_k, \sigma_k)_{k=1}^K \) maximizes the likelihood of the data \( (x_i)_{i=1}^N \) (equivalently, minimizes the negative-log-likelihood):
\begin{equation}
    \min_{(\alpha_k, \mu_k, \sigma_k)_{k=1}^K} \sum_{i=1}^N -\log\biggl( \sum_{k=1}^K \alpha_K G_{\mu_k,\sigma_k}(x_i) \biggr).
\end{equation}

The \gls{em} algorithm is a popular algorithm to solve this problem iteratively:
Given the estimates \( \alpha_k^{(j)}, \mu_k^{(j)}, \sigma_k^{(j)} \) at the \( j \)-th iteration, the expectation step computes the \emph{responsibilities} \( \gamma_{k, i}^{(j)} \) of the \( k \)-th component w.r.t\ the \( i \)-th data point as
\begin{equation}
	\gamma_{k, i}^{(j)} = \frac{\alpha_k^{(j)} w (x_i, \mu_k^{(j)}, \sigma_k^{(j)}) G_{\mu_k^{(j)}, \sigma_k^{(j)}}(x_i)}{\sum_{k=1}^K \alpha_k^{(j)} w (x_i, \mu_k^{(j)}, \sigma_k^{(j)}) G_{\mu_k^{(j)}, \sigma_k^{(j)}}(x_i)}.%
 	\label{eq:responsibilities}
\end{equation}
Later, we will show our choice of the weighting function \( w : \R^2 \times \R^2 \times \R_+ \to \R_+ \) to deal with outliers.
In the standard \gls{em} algorithm, \( w \equiv 1 \).

With the current responsibilities \( \gamma_{k, i}^{(j)} \), the maximization step amounts to updating
\begin{equation}
	\begin{aligned}
		\alpha_k^{(j+1)} &= \frac{1}{N} \sum_{i=1}^N \gamma_{k, i}^{(j)}, \\
		\mu_k^{(j+1)} &= \frac{\sum_{i=1}^N \gamma_{k,i}^{(j)} x_i}{\sum_{i=1}^N \gamma_{k, i}^{(j)}}, \\
		(\tilde\sigma_k^{(j+1)})^2 &= \frac{\sum_{i=1}^N \gamma_{k, i}^{(j)} \big\lVert x_i - \mu_k^{(j+1)}\big\rVert^2}{\sum_{i=1}^N \gamma_{k, i}^{(j)}}.
	\end{aligned}
\end{equation}
To avoid degenerate cases, we regularize the variances by adding a small positive scalar:
\begin{equation}
	\sigma_k^{(j + 1)} = \tilde\sigma_k^{(j + 1)} + \epsilon,
	\label{eq:sigma}
\end{equation}
where \( \epsilon > 0 \) is a tunable parameter, which should be set as small as possible while retaining stability.

In the fitting procedure, outliers present a major difficulty.
They are represented by keypoints that have been detected in only few of the warped images,
even just one,
and should not disturb the clustering process.
We tackle this by choosing \( w = w_1 \), where
\begin{equation}
    w_1(x, \mu, \sigma) = \begin{cases}
        1\ \text{if}\ \norm{x - \mu} < 3 \sigma, \\
        \exp\bigl( -(\norm{x - \mu} - 3\sigma^2)^2 / (2\sigma^2)\bigr)\ \text{else.}
    \end{cases}
\end{equation}%
Thus, points that are far from a component mean are adaptively down-weighted based on the distance to the center as well as the variance of the component.
To facilitate proper localization after the \gls{gmm} has been fit using the adapted \gls{em} algorithm, we run additional iterations ignoring points that are outside the \( 3\sigma \) radius, i.e. we set \( w = w_2 \), where
\begin{equation}
    w_2(x, \mu, \sigma) = \begin{cases}
        1 & \text{if}\ \norm{x - \mu} < 3 \sigma, \\
        0 & \text{else.}
    \end{cases}
\end{equation}

In addition, we perform a simple component selection scheme:
recall that the component means are initialized to the \gls{kde} maxima, \( \mathcal{X}_{\text{max}} \).
However, the \gls{kde} typically slightly overestimates the number of clusters; during the \gls{em} algorithm, components might \enquote{merge} to the same location.
In this case, the scores derived from the weights of the \gls{gmm} would systematically be too low as the weights are shared between two components.
To avoid this, we drop concentric components based on the distance of their means:
Let \( k, l \in \mathbb{N}, k < l\) be two component indices at iteration \( j \).
If \( \lVert \mu^{(j)}_{k} - \mu^{(j)}_{l} \rVert < \nu \), we discard the \( k \)-th component.
Here, \( \nu \in \R_+ \) defines the minimal distance between component means.

Note that our method is not necessarily restricted to isotropic covariance matrices;
we chose this as it makes the interpretation of the score straightforward.
An alternative would be to fit full covariance matrices and base the localization score on the largest singular value.

The resulting \emph{robustness} score for the k-th refined keypoint 
is given by the number of points within a \num{3}$\sigma$ radius of its gaussian component as:

\begin{equation}
    \emph{robustness}_k =  \sum_{i=1}^N w_2(x_i, \mu_k, \sigma_k),
\end{equation}

while the \emph{deviation} in pixels is defined as:

\begin{equation}
    \emph{deviation}_k = 6 \sigma_k. \\
\end{equation}

\section{Experiments}

\begin{table*}[t!]
    \def\mymidrule{%
        \cmidrule(r){1-1}\cmidrule(lr){2-4}\cmidrule(lr){5-7}\cmidrule(lr){8-10}\cmidrule(lr){11-13}\cmidrule(l){14-16}%
    }
    \setlength{\tabcolsep}{3 pt}
    \renewcommand{\arraystretch}{0.9}
    \centering
    \caption{Comparison on HPatches v-set - keypoints budget 2048.}%
    \begin{adjustbox}{width=\linewidth}
        \begin{tabular}{cc*{14}{S[table-format=1.3]}}
            \toprule
             & 
            \multicolumn{3}{c}{Rep @ $\uparrow$} & 
            \multicolumn{3}{c}{Rep MNN @ $\uparrow$} & 
            \multicolumn{3}{c}{MMA @ $\uparrow$} & 
            \multicolumn{3}{c}{MS @ $\uparrow$} &
            \multicolumn{3}{c}{Hom. Acc. AUC @ $\uparrow$} \\
             &
            {\qty{1}{\px}} & {\qty{2}{\px}} & {\qty{3}{\px}} & 
            {\qty{1}{\px}} & {\qty{2}{\px}} & {\qty{3}{\px}} & 
            {\qty{1}{\px}} & {\qty{2}{\px}} & {\qty{3}{\px}} & 
            {\qty{1}{\px}} & {\qty{2}{\px}} & {\qty{3}{\px}} & 
            {\qty{1}{\px}} & {\qty{3}{\px}} & {\qty{5}{\px}} \\
            \mymidrule
            Harris~\cite{harris} &
                    0.215 &     \bfseries 0.434 &     \bfseries 0.553 &  
                    0.215 &     0.398 &     0.458 &  
                    {---} &     {---} &     {---} &  
                    {---} &     {---} &     {---} &  
                    {---} &     {---} &     {---}  \\ 
            + GMM-IKRS &
                    \bfseries 0.229 &           0.426 &           0.532 &  
                    \bfseries 0.229 & \bfseries 0.421 & \bfseries 0.504 &  
                    {---} &     {---} &     {---} &  
                    {---} &     {---} &     {---} &  
                    {---} &     {---} &     {---}  \\ 
            \mymidrule
            DoG~\cite{sift} &
                    0.231 &          0.363 &      0.442 &  
                    0.231 &          0.362 &      0.428 &  
                    {---} &     {---} &     {---} &  
                    {---} &     {---} &     {---} &  
                    {---} &     {---} &     {---} \\ 
            + GMM-IKRS &
                    \bfseries 0.275 & \bfseries 0.433 & \bfseries 0.517 &  
                    \bfseries 0.275 & \bfseries 0.431 & \bfseries 0.502 &  
                    {---} &     {---} &     {---} &  
                    {---} &     {---} &     {---} &  
                    {---} &     {---} &     {---}  \\ 
            \mymidrule
            DISK~\cite{disk} &
                    0.211  &     0.388  &     0.464  &  
                    0.211  &     0.387  &     0.460  &  
                    0.352  &     0.621  &     0.709  & 
                    0.206  &     0.352  &     0.398 &  
                    0.092  &     0.299  &     0.420  \\ 
            + GMM-IKRS &
                \bfseries 0.257 & \bfseries 0.425 & \bfseries 0.499 &  
                \bfseries 0.257 & \bfseries 0.425 & \bfseries 0.494 &  
                \bfseries 0.409 & \bfseries 0.647 & \bfseries 0.727 & 
                \bfseries 0.248 & \bfseries 0.380 & \bfseries 0.422 &  
                \bfseries 0.110 & \bfseries 0.326 & \bfseries 0.444 \\ 
            \mymidrule
            SuperPoint~\cite{superpoint} &
                    0.197  &     0.392  &     0.502  &  
                    0.197  &     0.383  &     0.466  &  
                    0.242  &     0.483  &     0.595  &  
                    0.159  &     0.314  &     0.383  &  
                    0.121  &     0.360  &     0.488  \\ 
            + GMM-IKRS &
                \bfseries 0.247 & \bfseries 0.446 & \bfseries 0.545 &  
                \bfseries 0.247 & \bfseries 0.445 & \bfseries 0.538 &  
                \bfseries 0.312 & \bfseries 0.531 & \bfseries 0.627 &  
                \bfseries 0.211 & \bfseries 0.353 & \bfseries 0.413 &  
                \bfseries 0.144 & \bfseries 0.382 & \bfseries 0.507 \\ 
        \mymidrule
            R2D2~\cite{r2d2} &
                    0.163  &     0.354  &     0.459  &  
                    0.163  &     0.354  &     0.458  &  
                    0.261  &     0.520  &     0.625  &  
                    0.102  &     0.191  &     0.221  &  
                    0.069  &     0.253  &     0.367  \\ 
            + GMM-IKRS &
                \bfseries 0.205 & \bfseries 0.388 & \bfseries 0.486 &  
                \bfseries 0.205 & \bfseries 0.388 & \bfseries 0.484 &  
                \bfseries 0.308 & \bfseries 0.537 & \bfseries 0.629 &  
                \bfseries 0.121 & \bfseries 0.194 & \bfseries 0.222 &  
                \bfseries 0.078 & \bfseries 0.264 & \bfseries 0.376 \\ 
        \mymidrule
            MD-Net~\cite{mdnet} &
                    0.193 &    0.380 &     0.466 &  
                    0.193 &    0.378 &     0.462 &  
                    0.329 &    0.614 &     0.717 &  
                    0.167 &    0.296 &     0.339 &  
                    0.140 &    0.348 &     0.454\\ 
            + GMM-IKRS &
                \bfseries 0.237 & \bfseries 0.406 & \bfseries 0.486 &  
                \bfseries 0.237 & \bfseries 0.406 & \bfseries 0.483 &  
                \bfseries 0.388 & \bfseries 0.639 & \bfseries 0.729 &  
                \bfseries 0.199 & \bfseries 0.308 & \bfseries 0.345 &  
                \bfseries 0.143 & \bfseries 0.356 & \bfseries 0.470 \\ 
            \bottomrule
        \end{tabular}
    \end{adjustbox}
    \label{tab:hpatches}
\end{table*}

\subsection{Implementation Details}
We apply the same set of augmentations to each image:
\begin{itemize}
\item isotropic scaling \{\num{1.5}, \num{1.25}, \num{0.75}, \num{0.5}\},
\item anisotropic scaling along $x$, $y$ \{\num{1.5}, \num{1.25}, \num{0.75}, \num{0.5}\},
\item anisotropic shear along $x$, $y$ \{\num[explicit-sign=+]{+0.2}, \num{-0.2}, \num[explicit-sign=+]{+0.6}, \num{-0.6}\}.
\end{itemize}
These result in a set of 21 images, including the original one.
We found beneficial to add a slight Gaussian noise to all the warped images.
This prevents weak noise patterns in the original image from triggering repeated detections.

In all the experiments we extract \( N_{\text{kpts}} = \num{2048} \) keypoints from each image.
For the \gls{kde}, we choose the bandwidth \( h = \qty{0.5}{\px} \) and a window radius of \( r = \qty{3}{\px} \) to identify the maxima \( \mathcal{X}_\text{max} \).
We initialize the \gls{gmm} component means with the entries of \( \mathcal{X}_\text{max} \) and choose the initial variance such that the diameter at \( 3\sigma \) corresponds to \qty{2}{\px} and set \( \nu = \qty{0.1}{\px} \).
At each iteration, we also clamp the maximum variance for each component
such that its diameter at \( 3\sigma \) does not exceed \qty{10}{\px}.
In addition, 
we consider the rare case that clusters may contain more than one keypoint extracted from the same image. 
This may happen when two badly localized clusters of keypoints partially intersect and the
\gls{gmm} uses a single component to fit both.
In this case, we correct the cluster weight by removing the contributions from the duplicate keypoints.

For all the deep methods, 
we sample the descriptors from their dense descriptor volume at the integer locations of the refined keypoints.
We run our pipeline with exactly the same parameters regardless of the keypoint detector used,
which confirms the generality of our process.

\begingroup
\begin{table*}[t!]
\centering
\def\mymidruleA{%
    \cmidrule(lr){2-4}\cmidrule(l){5-7}%
}
\def\mymidruleB{%
    \cmidrule(r){1-1}\cmidrule(lr){2-4}\cmidrule(l){5-7}%
}
\caption{Comparison on IMC~\cite{imb} stereo, keypoints budget 2048.}
\begin{adjustbox}{width=0.9\linewidth}
\setlength{\tabcolsep}{4 pt}
\renewcommand{\arraystretch}{1.0}
\centering
\begin{tabular}{c*{2}{S[table-format=0.2]S[table-format=3]S[table-format=0.3]}}
\toprule
    & 
    \multicolumn{3}{c}{Validation set} &
    \multicolumn{3}{c}{Test set} \\
     & 
    {Rep@\qty{3}{\px} $\uparrow$} & 
    {N. inliers $\uparrow$} & 
    {mAA@10° $\uparrow$} &
    {Rep@\qty{3}{\px} $\uparrow$} & 
    {N. inliers $\uparrow$} & 
    {mAA@10° $\uparrow$}  \\
\mymidruleB
    DISK~\cite{disk} &
        0.493 & 
        438 &
        0.707 &
        0.452 &
        393 &
        0.503  \\
    + GMM-IKRS &
        \bfseries 0.507 & 
        \bfseries 469 &
        \bfseries 0.722 &
        \bfseries 0.468 &
        \bfseries 426 &
        \bfseries 0.524 \\
\mymidruleB
    SuperPoint~\cite{superpoint} &
        0.350 & 
        154 &
        0.270 &
        0.359 &
        176 &
        0.282 \\
    + GMM-IKRS &
        \bfseries 0.380 & 
        \bfseries 182 &
        \bfseries 0.319 &
        \bfseries 0.391 &
        \bfseries 210 &
        \bfseries 0.318 \\
\mymidruleB
    R2D2~\cite{r2d2} &
        0.377 & 
        117 &
        0.307 &
        0.388 &
        134 &
        0.271 \\
    + GMM-IKRS &
        \bfseries 0.396 & 
        \bfseries 120 &
        \bfseries 0.313 &
        \bfseries 0.406 &
        \bfseries 135 &
        \bfseries 0.276 \\
\mymidruleB
    MD-Net~\cite{mdnet} &
        0.357 & 
        144 &
        0.470 &
        0.396 &
        199 &
        0.401 \\
    + GMM-IKRS &
        \bfseries 0.370 & 
        \bfseries 152 &
        \bfseries 0.484 &
        \bfseries 0.411 &
        \bfseries 211 &
        \bf 0.412 \\
        
\bottomrule
\end{tabular}
\end{adjustbox}
\label{tab:imb}
\end{table*}
\endgroup

\begingroup
\begin{table*}[t!]
\centering
\def\mymidruleA{%
    \cmidrule(lr){2-5}\cmidrule(l){6-9}%
}
\def\mymidruleB{%
    \cmidrule(r){1-1}\cmidrule(lr){2-5}\cmidrule(l){6-9}%
}
\caption{Comparison on IMC~\cite{imb} multiview, keypoints budget 2048.}
\begin{adjustbox}{width=0.9\linewidth}
\setlength{\tabcolsep}{4 pt}
\renewcommand{\arraystretch}{1.0}
\centering
\begin{tabular}{c*{2}{S[table-format=4]S[table-format=0.3]S[table-format=3]S[table-format=0.3]}}
\toprule
    & 
    \multicolumn{4}{c}{Validation set} &
    \multicolumn{4}{c}{Test set} \\
     & 
    {3D pts. $\uparrow$} & 
    {Track L. $\uparrow$} & 
    {N. Inliers $\uparrow$} &
    {mAA@10° $\uparrow$}  &
    {3D pts. $\uparrow$} & 
    {Track L. $\uparrow$} & 
    {N. Inliers $\uparrow$} &
    {mAA@10° $\uparrow$}  \\
\mymidruleB
    DISK~\cite{disk} &
        2195 & 
        6.032 &
        450 &
        0.842 &
        2183 &
        5.660 &
        404 &
        0.729 \\
    + GMM-IKRS &
        \bfseries 2217 & 
        \bfseries 6.088 &
        \bfseries 482 &
        \bfseries 0.850 &
        \bfseries 2203 &
        \bfseries 5.736 &
        \bfseries 438 &
        \bfseries 0.736 \\
\mymidruleB
    SuperPoint~\cite{superpoint} &
        1332 & 
        4.330 &
        163 &
        0.568 &
        1522 &
        4.356 &
        181 &
        0.596 \\
    + GMM-IKRS &
        \bfseries 1467 & 
        \bfseries 4.488 &
        \bfseries 193 &
        \bfseries 0.601 &
        \bfseries 1685 &
        \bfseries 4.466 &
        \bfseries 217 &
        \bfseries 0.623 \\
\mymidruleB
    R2D2~\cite{r2d2} &
        1022 & 
        4.702 &
        127 &
        0.547 &
        1287 &
        4.462 &
        139 &
        0.548 \\
    + GMM-IKRS &
        \bfseries 1044 & 
        \bfseries 4.710 &
        \bfseries 131 &
        \bfseries 0.556 &
        \bfseries 1309 &
        \bfseries 4.473 &
        \bfseries 141 &
        \bfseries 0.560 \\
\mymidruleB
    MD-Net~\cite{mdnet} &
        1108 & 
        5.002 &
        154 &
        0.727 &
        1474 &
        5.020 &
        205 &
        0.692 \\
    + GMM-IKRS &
        \bfseries 1133 & 
        \bfseries 5.053 &
        \bfseries 162 &
        \bfseries 0.740 &
        \bfseries 1511 &
        \bfseries 5.072 &
        \bfseries 216 &
        \bfseries 0.704 \\
        
\bottomrule
\end{tabular}
\end{adjustbox}
\label{tab:imb_multiview}
\end{table*}
\endgroup

\subsection{HPatches}
\label{sec:hpatches}
We test our framework on the popular HPatches~\cite{hpatches} dataset.
We run GMM-IKRS on top of several recent deep local feature extractors
as well as hand-crafted classical keypoint detectors which have been widely used in the past.
To test the performances of our method when dealing with viewpoint changes, 
we focus our experiments on the \textit{v} set of HPatches, 
which contains photos of planar scenes taken from different viewpoints.
Each scene contains \qty{1} reference image and \qty{5} source images of different viewpoints. 
We run all the methods using the code provided by the authors and use
OpenCV~\cite{opencv} RANSAC with \qty{3.0}{\px} threshold and \num{100}k iterations to recover the homographies.
As often done in the literature, we extract a maximum of \num{2048} keypoints from each image
and match the descriptors using a simple \gls{mnn} search, 
without any minimum score or ratio test.
The results of our evaluations are reported in Table~\ref{tab:hpatches}, 
where we compare the performance of each method with or without our refinement
for various metrics at different pixel thresholds:
\begin{itemize}
\item \tbf{Repeatability}: 
Ratio between the number of keypoints that, from any of the two images,
project close to a keypoint in the other and the total number of keypoints in the area of overlap.
\item \tbf{Repeatability-Mutual-Nearest-Neighbor (MNN)}: 
This is a metric we propose to compensate for the overestimates of the standard repeatability metric in case of dense keypoint detections.
A keypoint from image A is considered MNN-repeatable, if it is both repeatable and forms a mutual-nearest-neighbor pair in both images with the same keypoint from image B.
The repeatability-MNN metric is then computed as the ratio between the sum of the number of mnn-repeatable keypoints 
from the two images and the total number of keypoints in the overlap area.
This metric better relates to the pairwise descriptor matching task and upper-bounds to the standard repeatability in the case of well spaced keypoints.
\item \tbf{Mean Matching Accuracy}: Ratio between the number of correct matches and the number of proposed matches.
\item \tbf{Matching Score}: Ratio between the number of correct matches and the average number of detected keypoints in the overlap area.
\item \tbf{Homography Accuracy AUC}: Area under the curve of the fraction of image pairs where the relative homography could be recovered 
with an average corner error lower than a specified threshold, evaluated at 0.1px steps.
\end{itemize}
The table shows a consistent improvement, with a single exception, over all the computed metrics at all pixel thresholds.
Harris~\cite{harris}, due to its densely detected keypoints, tricks the repeatability measure obtaining higher scores at \num{2} and \num{3} pixels,
but falls behind once evaluated with the more robust repeatability-MNN metric.
The Homography Accuracy AUC,
which is the most important score for many downstream tasks,
is also improved when using our refined keypoints.
The performance boost obtained using 
the refined keypoints
does not only come from the sub-pixel accuracy of our method,
as better discussed in Sec~\ref{sec:ablation_studies}, but also from a more robust selection of keypoints. 
This is confirmed by the improved repeatabilities obtained for the already sub-pixel-accurate DoG~\cite{sift} detector.

\begin{figure*}[t]
     \newcommand{\mysize}{0.20\linewidth}
     \centering
     \begin{subfigure}[b]{\mysize}
         \includegraphics[width=\linewidth, trim={0.0cm 3.5cm 0.0cm 2.0cm}, clip]{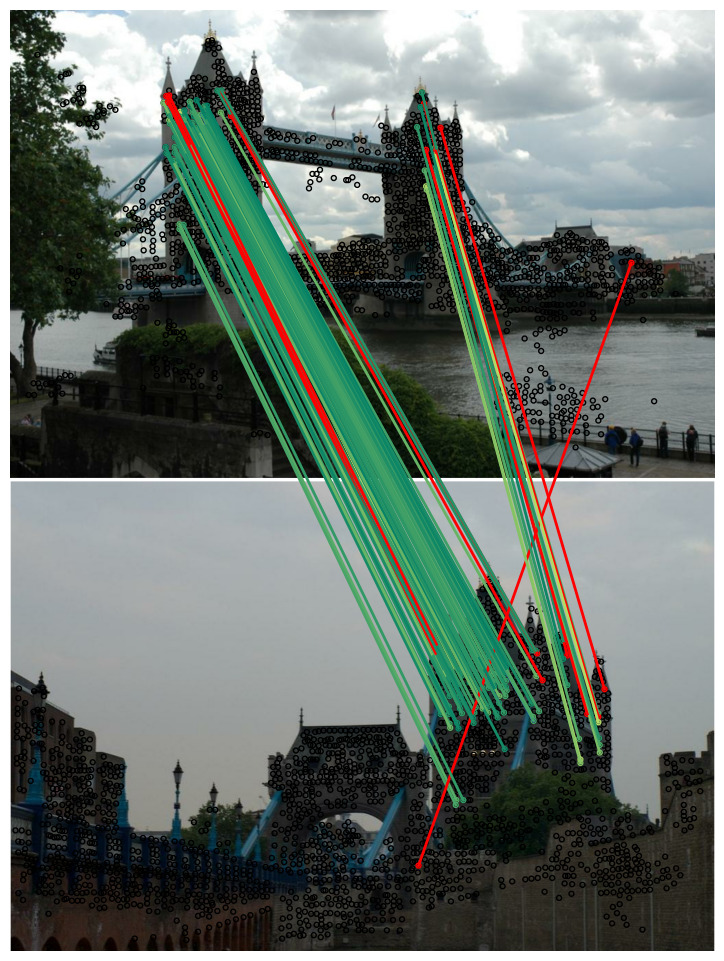}
     \end{subfigure}
     \begin{subfigure}[b]{\mysize}
         \includegraphics[width=\linewidth, trim={0.0cm 3.5cm 0.0cm 2.0cm}, clip]{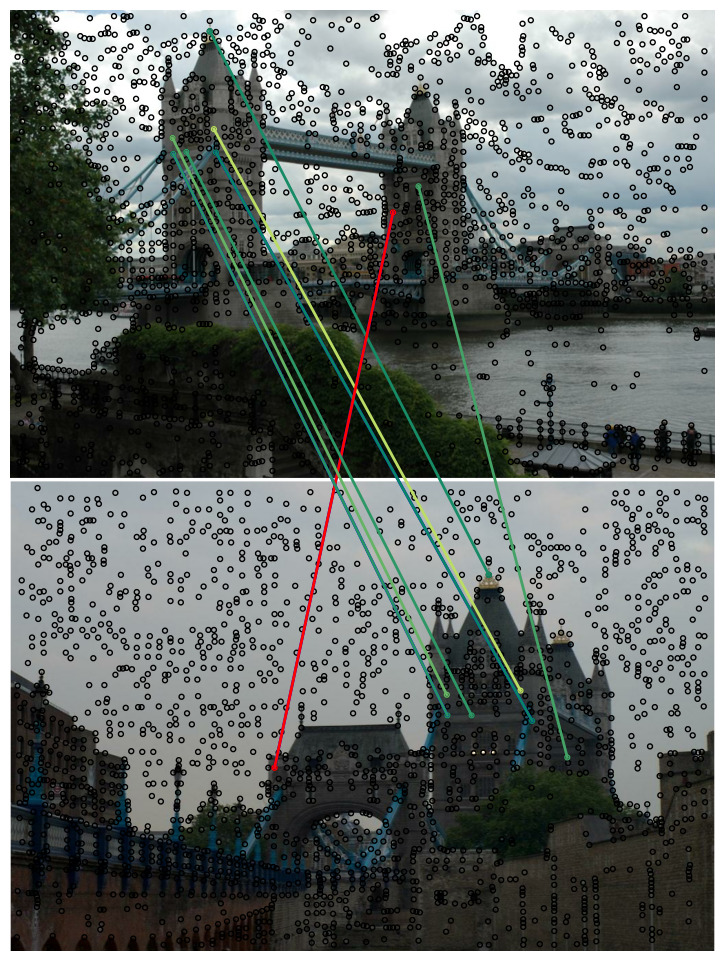}
     \end{subfigure}
     \begin{subfigure}[b]{\mysize}
         \includegraphics[width=\linewidth, trim={0.0cm 3.5cm 0.0cm 2.0cm}, clip]{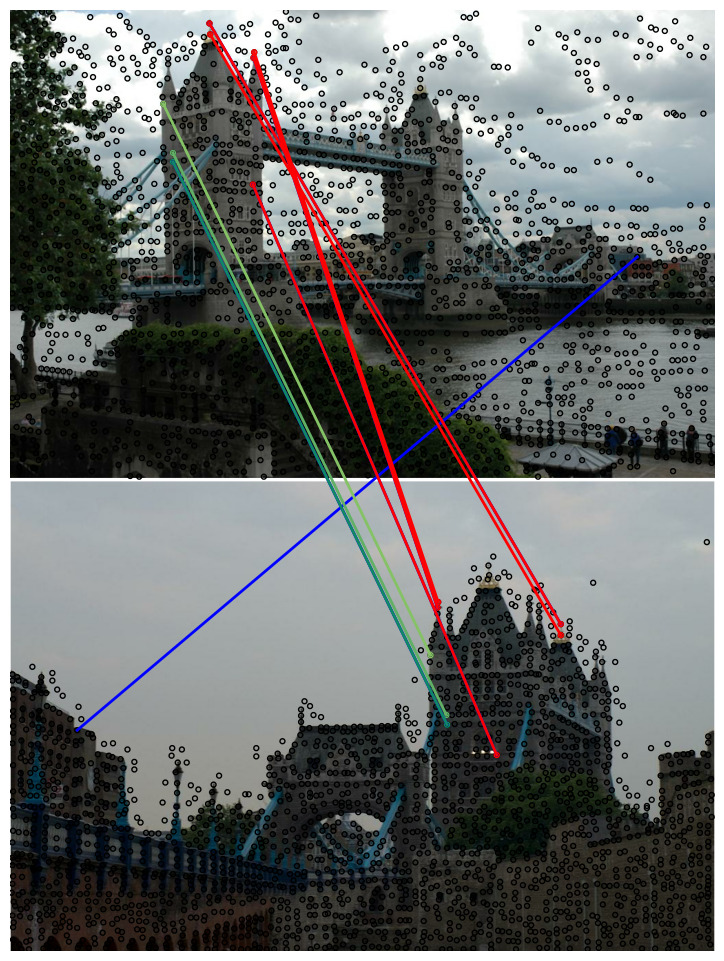}
     \end{subfigure}
     \begin{subfigure}[b]{\mysize}
         \includegraphics[width=\linewidth, trim={0.0cm 3.5cm 0.0cm 2.0cm}, clip]{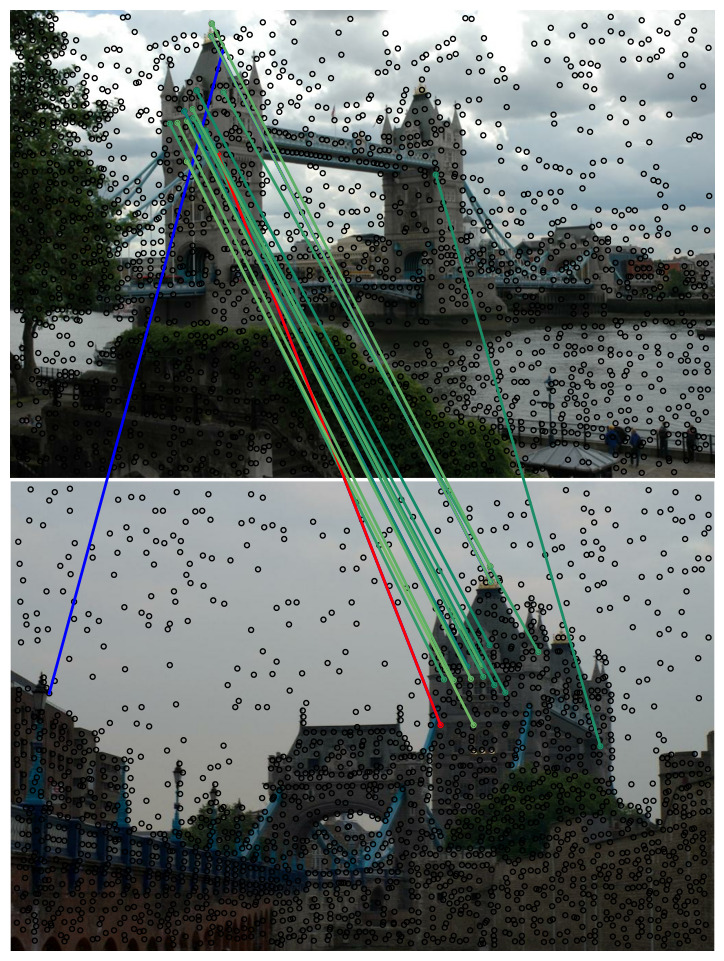}
     \end{subfigure}
     \\
     \begin{subfigure}[b]{\mysize}
         \includegraphics[width=\linewidth, trim={0.0cm 3.5cm 0.0cm 2.0cm}, clip]{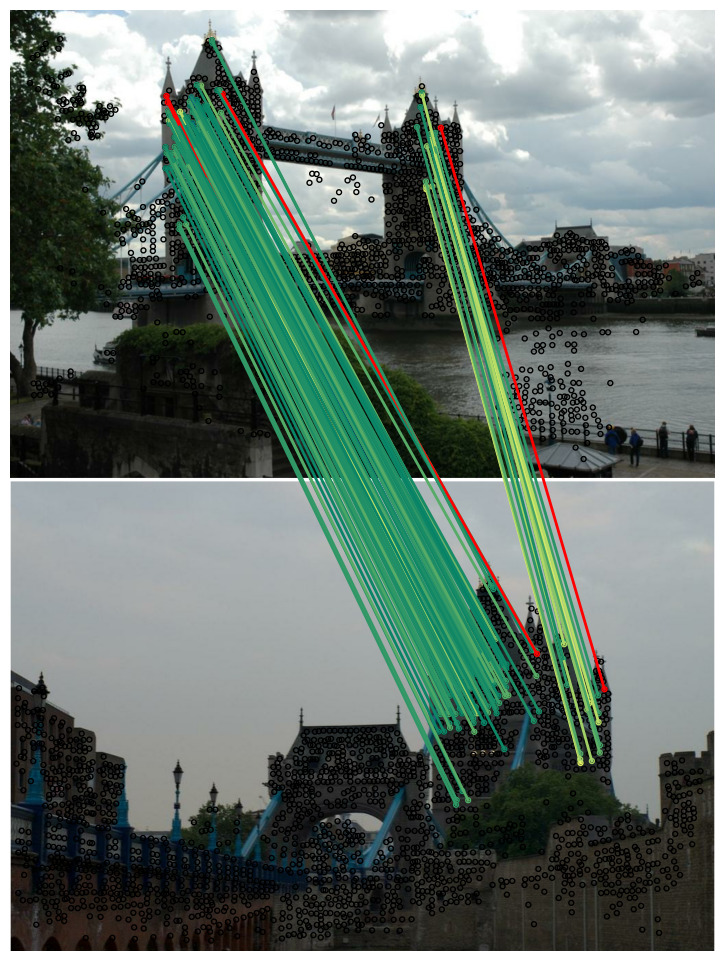}
         \caption{DISK~\cite{disk}}
     \end{subfigure}
     \begin{subfigure}[b]{\mysize}
         \includegraphics[width=\linewidth, trim={0.0cm 3.5cm 0.0cm 2.0cm}, clip]{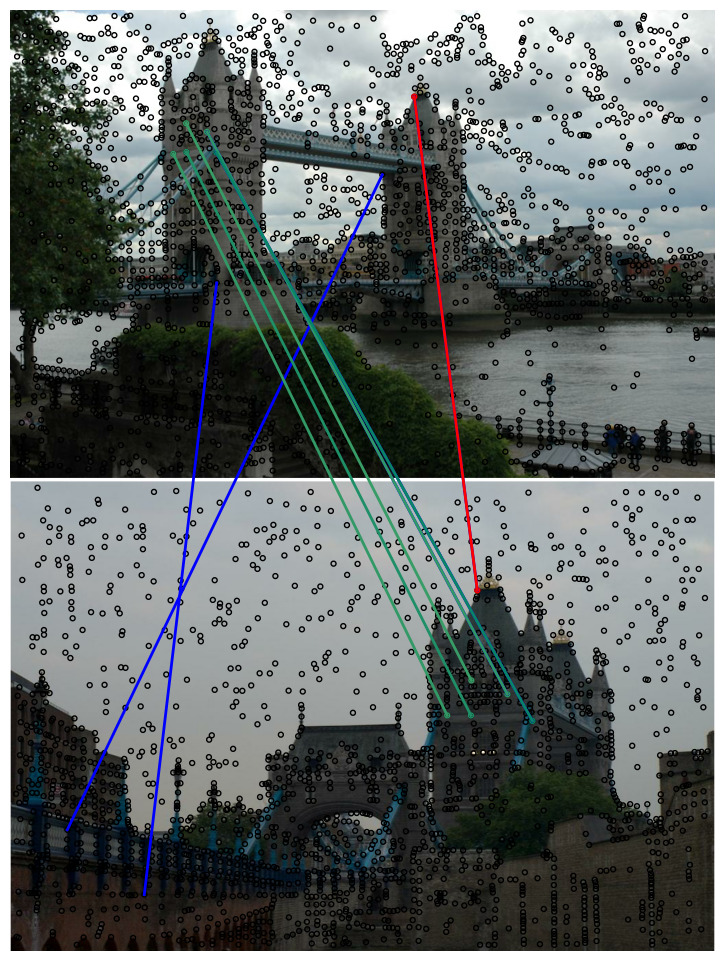}
         \caption{SuperPoint~\cite{superpoint}}
     \end{subfigure}
     \begin{subfigure}[b]{\mysize}
         \includegraphics[width=\linewidth, trim={0.0cm 3.5cm 0.0cm 2.0cm}, clip]{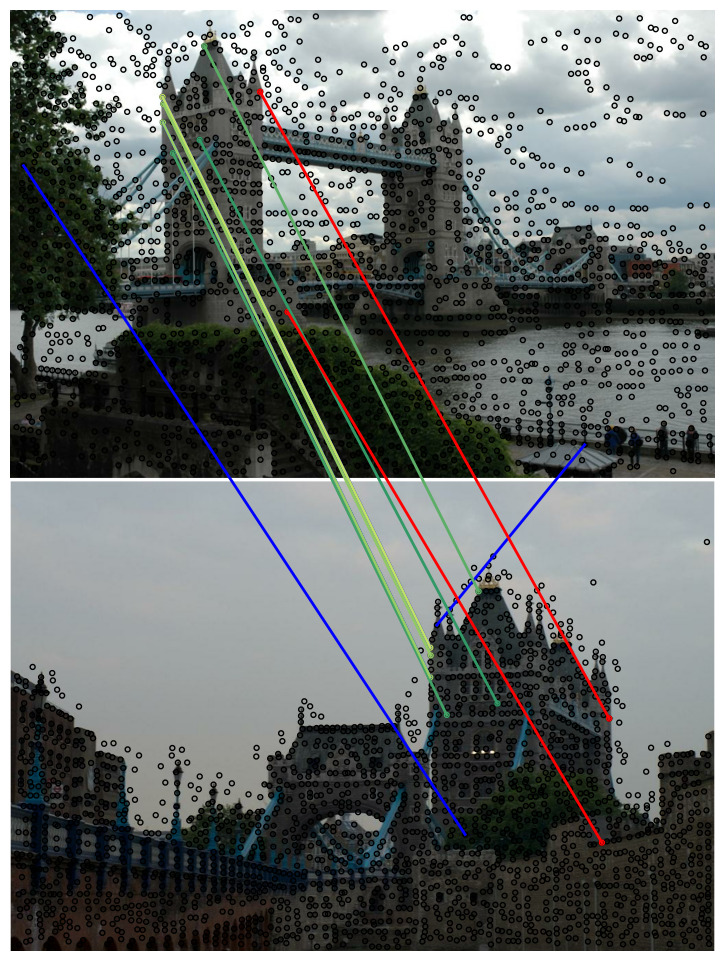}
         \caption{R2D2~\cite{r2d2}}
     \end{subfigure}
     \begin{subfigure}[b]{\mysize}
         \includegraphics[width=\linewidth, trim={0.0cm 3.5cm 0.0cm 2.0cm}, clip]{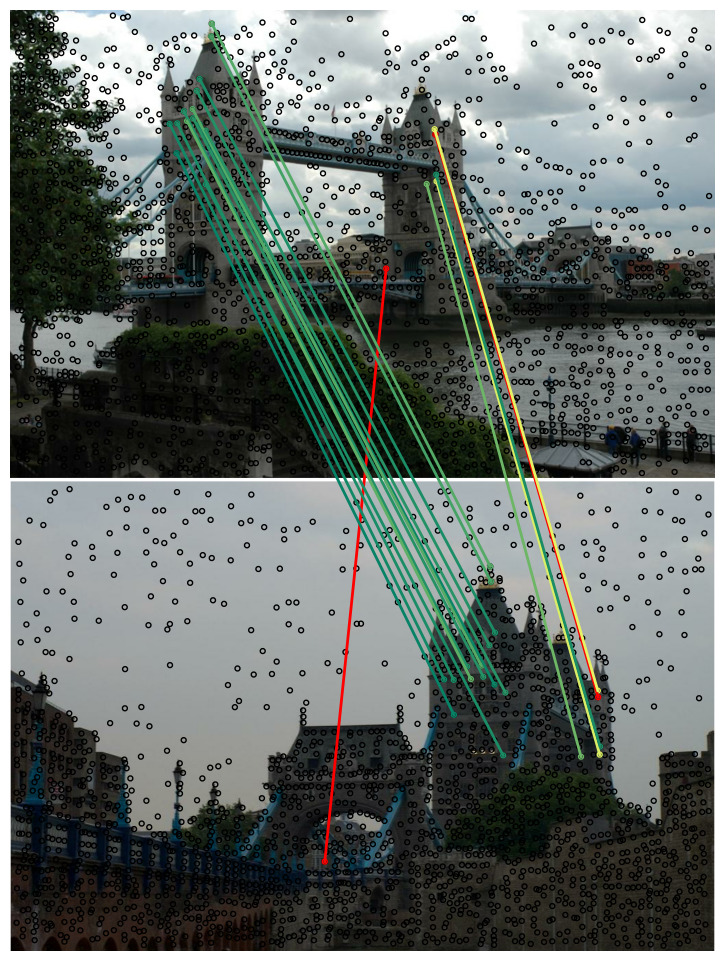}
         \caption{MD-Net~\cite{mdnet}}
     \end{subfigure}
     \caption{
     Visual comparison between original methods (top) and their GMM-IKRS refined version (bottom) on a challenging image pair from Image Matching Challenge~\cite{imb}.
     The RANSAC inliers are color coded depending on the reprojection error, from green (0px) to yellow (5px). 
     Matches are shown in red when their reprojection error is larger than \qty{5}{px}, and in blue when the depth is not available. 
     For this image pair, all the methods except SuperPoint~\cite{superpoint} obtain more and better localized matches.
     }
     \label{fig:imb}
\end{figure*}

\begin{table}[t]
\centering
\captionof{table}{Ablations on HPatches v-set.}
\begingroup
\setlength{\tabcolsep}{4 pt}
\renewcommand{\arraystretch}{1.2}
\begin{adjustbox}{width=0.95\linewidth}
\centering
\begin{tabular}{cccccc}
\toprule
    {Method} & 
    {Rep@\qty{3}{\px}$\uparrow$} & 
    {Rep MNN@\qty{3}{\px}$\uparrow$} & 
    {MMA@\qty{3}{\px} $\uparrow$} &
    {MS@\qty{3}{\px} $\uparrow$} &
    {s per img} \\
\hline
    {DISK}~\cite{disk} &
     0.464 &
     0.460 &
     0.709 &
     0.398 &
     0.03 \\
    + 8 {warps} &
     0.484 &
     0.479 &
     0.722 &
     0.412 &
     0.43 \\
    + 14 {warps} &
     0.493 &
     0.488 &
     \bfseries 0.732 &
     0.416 &
     0.69 \\
    \bfseries + 20 {warps} &
    \bfseries 0.499 &
    \bfseries 0.494 &
     0.727 &
    \bfseries 0.422 &
     0.98 \\
\hline
    + 20 {warps w/o outlier rejection} &
     0.483 &
     0.478 &
     0.724 &
     0.395 &
     0.98 \\
    + 20 {warps KDE only} &
     0.473 &
     0.468 &
     0.719 &
     0.398 &
     0.78 \\
    + 20 {warps round int} &
     0.494 &
     0.489 &
     0.726 &
     0.417 &
     0.98 \\
    {DISK}~\cite{disk} + 20 heatmap warps &
     0.488 &
     0.482 &
     0.726 &
     0.413 &
     0.65 \\

\bottomrule
\end{tabular}
\end{adjustbox}
\label{tab:ablations}
\endgroup
\end{table}

\subsection{Image matching benchmark}
To evaluate the generalization of our framework to more challenging scenarios,
we run the deep learning based methods, with and without our refinement framework, 
on the phototourism set of the 2021 Image Matching Challenge (IMC)~\cite{imb}.
Additionally to the stereo pose recovery task, 
the benchmark evaluates the local features performance in the more practical scenario of multi-view 3D reconstruction.
Each of the scenes (\num{3} validation set, \num{9} test set) contains \num{100} pictures
of famous landmarks, captured by tourist with different cameras, 
from various angles and under diverse lighting conditions
We run each method single-scale, using the matching and filtering parameters obtained from the online leaderboard, where available.
The numerical results are shown in Table~\ref{tab:imb} and Table~\ref{tab:imb_multiview}, 
where the methods are evaluated for:
\begin{itemize}
    \item \tbf{mAA}: Mean Average Accuracy. This is the main benchmark metric, computed as the area under the curve of the fraction of relative poses recovered within a maximum error in degrees,
        evaluated at \num{1}° steps.
    \item \tbf{N. inliers}: Average number of inlier matches after the robust fit.
    \item \tbf{3D pts.}: Average number of reconstructed 3D points.
    \item \tbf{Track L.}: Average 3D reconstruction track length.
\end{itemize}
For all methods, every metric improves,
thus confirming the validity of our refinement approach for real tasks.
SuperPoint~\cite{superpoint}, with a more than 10\% improvement in stereo mAA and 7\% in multiview mAA, 
is the method that most strongly benefits from our framework, 
followed by DISK~\cite{disk} and MD-Net~\cite{mdnet}.

\begin{figure*}[t!]
\centering
\begin{subfigure}[b]{0.49\textwidth}
\centering
\begin{tikzpicture}
\begin{scope}[scale=0.7]
	\begin{groupplot}[
		group style={
			group name=myplot,
			group size= 1 by 4,
			x descriptions at=edge bottom,
			vertical sep=0pt,
		},
		height=4cm,
		width=9.5cm,
        cycle list name=colorbrewer-RYB,
	]
			\nextgroupplot[bar width=6pt]
				\foreach \method in {HarrisCorners, DoG, DISK, SuperPoint, R2D2, MD-Net} {
					\addplot+[mark size=1pt,ybar,opacity=.5] table [col sep=comma,] {./media/robustness/1/\method.csv};
				}
			\nextgroupplot[bar width=6pt,scaled y ticks=manual:{}{\pgfmathparse{#1/10000}}]
				\foreach \method in {HarrisCorners, DoG, DISK, SuperPoint, R2D2, MD-Net} {
					\addplot+[mark size=1pt,ybar,opacity=.5] table [col sep=comma,] {./media/robustness/2/\method.csv};
				}
			\nextgroupplot[bar width=6pt,scaled y ticks=manual:{}{\pgfmathparse{#1/10000}}]
				\foreach \method in {HarrisCorners, DoG, DISK, SuperPoint, R2D2, MD-Net} {
					\addplot+[mark size=1pt,ybar,opacity=.5] table [col sep=comma,] {./media/robustness/3/\method.csv};
				}
			\nextgroupplot[bar width=6pt,scaled y ticks=manual:{}{\pgfmathparse{#1/10000}}]
				\foreach \method in {HarrisCorners, DoG, DISK, SuperPoint, R2D2, MD-Net} {
					\addplot+[mark size=1pt,ybar,opacity=.5] table [col sep=comma,] {./media/robustness/999/\method.csv};
				}
	\end{groupplot}
    \node [rotate=90] at (-.4, -2.5) {\tiny Num points};
    \node at (3.5, -7.7) {\tiny robustness};
\end{scope}
\end{tikzpicture}
\caption{Cluster \emph{robustness} distribution at different \emph{deviation} thresholds. 
From top to bottom, the maximum \emph{deviation} is set to \qty{1}{\px}, \qty{2}{\px}, \qty{3}{\px}, \hinfty.}
\label{fig:hist_weight}
\end{subfigure}\hfill
\begin{subfigure}[b]{0.49\textwidth}
    \centering
\begin{tikzpicture}
\begin{scope}[scale=0.7]
	\begin{groupplot}[
		group style={
			group name=myplot,
			group size= 1 by 4,
			x descriptions at=edge bottom,
			vertical sep=0pt,
		},
		height=4cm,
		width=9.5cm,
        cycle list name=colorbrewer-RYB,
        legend style={nodes={scale=0.5, transform shape}},
	]
			\nextgroupplot[bar width=6pt, legend to name=legend]
				\foreach \method in {HarrisCorners, DoG, DISK, SuperPoint, R2D2, MD-Net} {
					\addplot+[mark size=1pt, ybar, opacity=.5] table [col sep=comma,] {./media/deviation/5/\method.csv};
                    \expandafter\addlegendentry\expandafter{\method}
            }
			\nextgroupplot[bar width=6pt,scaled y ticks=manual:{}{\pgfmathparse{#1/10000}}]
				\foreach \method in {HarrisCorners, DoG, DISK, SuperPoint, R2D2, MD-Net} {
					\addplot+[mark size=1pt, ybar, opacity=.5] table [col sep=comma,] {./media/deviation/10/\method.csv};
				}
			\nextgroupplot[bar width=6pt,scaled y ticks=manual:{}{\pgfmathparse{#1/10000}}]
				\foreach \method in {HarrisCorners, DoG, DISK, SuperPoint, R2D2, MD-Net} {
					\addplot+[mark size=1pt, ybar, opacity=.5] table [col sep=comma,] {./media/deviation/15/\method.csv};
				}
			\nextgroupplot[bar width=6pt,scaled y ticks=manual:{}{\pgfmathparse{#1/10000}}]
				\foreach \method in {HarrisCorners, DoG, DISK, SuperPoint, R2D2, MD-Net} {
					\addplot+[mark size=1pt, ybar, opacity=.5] table [col sep=comma,] {./media/deviation/20/\method.csv};
				}
	\end{groupplot}
        \node[below] at (6, 0.25) {\pgfplotslegendfromname{legend}};
        \node [rotate=90] at (-.5, -2.5) {\tiny Num points};
        \node at (3.5, -7.7) {\tiny deviation};
\end{scope}
\end{tikzpicture}
    \caption{Cluster \emph{deviation} distribution at different \emph{robustness}. 
    From top to bottom, the minimum \emph{robustness} is set to \num{5}, \num{10}, \num{15}, \num{20} (max is \num{21}).}
    \label{fig:hist_sigma}
\end{subfigure}
\caption{Statistical analysis of keypoints extracted from different detectors on the HPatches\cite{hpatches} v-set images. Better seen in color. }
\label{fig:statistics}
\end{figure*}
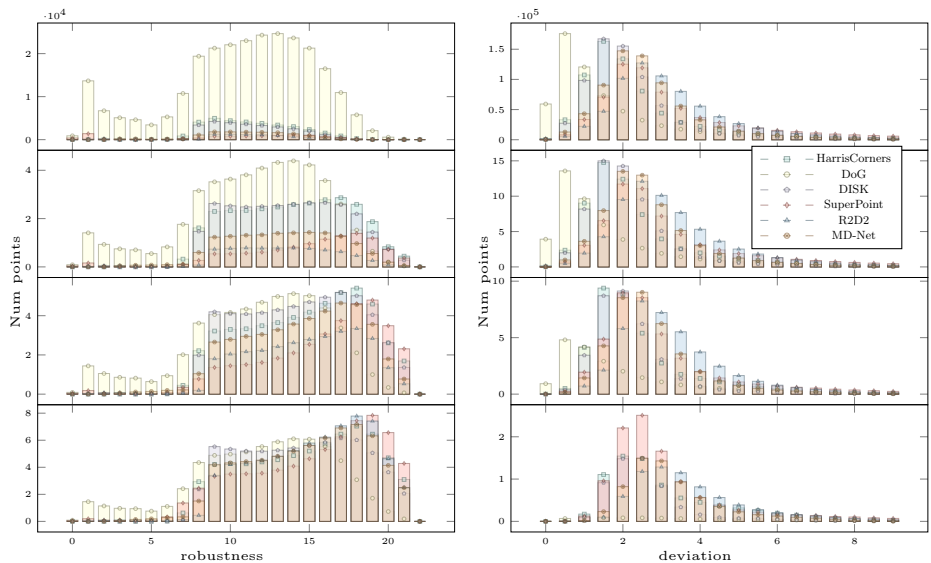

\subsection{Statistical analysis}
In addition to the
metrics evaluated for the benchmark,
our framework permits to shed more light on the different detector behaviours.
In Figure~\ref{fig:statistics}, we report the
distributions of \emph{robustness} and \emph{deviation} 
for all the keypoints extracted from HPatches.
We recall that a \emph{deviation} value of \qty{1}{\px} indicates a cluster whose diameter at \num{3}$\sigma$ measures \qty{1}{\px}.
From the histograms we can notice how DoG~\cite{sift}, 
which obtains similar repeatabilities to other methods on the HPatches evaluations,
is characterized by very different \emph{robustness} and \emph{deviation} distributions.
Looking at the top-left chart, which shows the \emph{robustness} distribution for
well localized keypoints with \emph{deviation} up to \qty{1}{\px},
it can be noticed how DoG dominates all the other methods;
not only at higher \emph{robustness} values, but also at lower ones.
This leads to the conclusion that DoG is very good at finding well localized keypoints,
but it does not excel at finding robust ones.
When relaxing the requirements for localization accuracy to \qty{3}{\px},
shown in the 3rd row, methods like DISK, Harris and MD-Net
are able to detect keypoints with higher \emph{robustness} scores,
that is points that are very likely to be detected again in another image.
In the last row of the same chart, 
SuperPoint emerges as the method able to find the largest number of very robust keypoints.
The distributions on the right chart describe instead how well localized
keypoints with different degrees of robustness are.
If looking for very robust keypoints, we should focus
on the bottom chart, which shows the distributions of keypoints with \emph{robustness} score of at least \num{20}/\num{21}. 
In this case, DoG results the worst performing method, 
whereas SuperPoint is the one able to find the largest number of keypoints
with \emph{deviation} lower than \qty{4}{px}.

In Figure~\ref{fig:best_keypoints} 
we show for each method the best refined keypoint across 5 different HPatches scenes.
We can notice a tendency of classical method to prefer hard contrast regions.
SuperPoint, in accordance with its specific pretraining,
shows a clear preference for corner-like structures, while DISK, R2D2 and MD-Net 
seem more oriented toward blobs.
In particular, 
DISK appears able to find robust keypoints even in less contrasted regions (rows 2 and 5).

\section{Ablation studies}
\label{sec:ablation_studies}
To validate our design choices, we conduct further experiments using DISK~\cite{disk},
which was found to be the best performing method on IMC~\cite{imb}.
The results are shown in Table~\ref{tab:ablations}.
On the upper section of the table, we compare our framework performance using different numbers of augmentations.
In the \num[explicit-sign=+]{+8} \textit{warps} row, we use only the smallest shears and anisotropic scalings from the set of augmentations, while
in the \num[explicit-sign=+]{+14} \textit{warps} row we reintroduce the isotropic scaling and stronger shears.
When comparing results with different augmentation numbers,
we can observe that the \num[explicit-sign=+]{+20} \textit{warps} row yields the best trade off in term of performances,
keeping the computational time below one second.
The mean keypoint refinement shift in this case is \qty{0.34}{\px}.
In the \textit{w/o outlier rejection} line we report the results obtained dropping our robustified GMM modification and, for
the \textit{warps KDE only} entry, we directly use the \gls{kde} local maxima as keypoints, skipping the \gls{gmm} fit.
For a more fair comparison against the discrete-pixel-accurate \gls{kde} result, 
we also report the results obtained rounding our GMM-IKRS keypoints coordinate to the closest integer.
Comparing against these lines, we can see how our robust \gls{gmm}-fit scores and refines the keypoint locations 
better and beyond a simple sub-pixel refinement.
Finally, as a baseline comparison, we report as \textit{DISK \num[explicit-sign=+]{+20} heatmap warps} the results
obtained aggregating the unwarped
DISK heatmaps directly and then detecting the keypoints.
This approach improves the original DISK results, but not as much as our GMM-IKRS pipeline.

\section{Conclusion}
In this work, we presented GMM-IKRS,
a general framework capable of refining and evaluating the keypoints from any detector.
The two scores assigned by our pipeline, \emph{robustness} and \emph{deviation},
characterize each keypoint in an interpretable manner and
can be compared across different methods.
This permits an in-depth analysis of qualities across different detectors,
which would otherwise not be possible.
The outcome of our experiments on the HPatches v-set confirmed the validity of our method, 
while the results of the Image Matching Challenge
demonstrated its refinement capabilities for 3D reconstruction, even under challenging conditions.
For the future, an interesting research direction could consist in the use of our method
for the generation of sub-pixel keypoints to be used as ground truth to train, in a teacher-student fashion, a deep detector.

\clearpage 

\section*{Acknowledgement}
This work has been supported by the FFG, Contract No. 881844: "Pro²Future is funded within the Austrian COMET Program 
Competence Centers for Excellent Technologies under the auspices of the Austrian Federal Ministry for Climate Action, 
Environment, Energy, Mobility, Innovation and Technology, the Austrian Federal Ministry for Digital and 
Economic Affairs and of the Provinces of Upper Austria and Styria. COMET is managed by the Austrian Research Promotion 
Agency FFG."

\bibliographystyle{splncs04}
\bibliography{main}
\end{document}